\documentclass{article}

\usepackage{arxiv}

\usepackage{lineno} 

\usepackage{times}
\usepackage{latexsym}

\usepackage[T1]{fontenc}

\usepackage[utf8]{inputenc}

\usepackage{microtype}

\usepackage{inconsolata}
\usepackage{amsmath}
\usepackage{graphicx} 
\usepackage{dsfont}
\usepackage{enumitem}
\usepackage{pgfplots}
\usepackage{subcaption}
\usepackage{booktabs}
\usepackage{float}
\usepackage{xspace}
\usepackage{multirow}

\usepackage{balance}
\pgfplotsset{compat=newest}
\usepackage[listings,breakable,skins]{tcolorbox}
\usepackage{refcount}
\usepackage{stfloats} 
\usepackage{tablefootnote}
\usepackage{url}
\usepackage{nameref}
\usepackage{authblk}

\newcommand{\Comment}[1]{}

\begin{document}

\newcommand{\idf}{{\em IDF}}
\newcommand{\meanIDF}{{\em meanIDF}}
\newcommand{\maxIDF}{{\em maxIDF}}
\newcommand{\minIDF}{{\em minIDF}}
\newcommand{\scq}{{\em SCQ}}
\newcommand{\meanSCQ}{{\em meanSCQ}}
\newcommand{\maxSCQ}{{\em maxSCQ}}
\newcommand{\minSCQ}{{\em minSCQ}}
\newcommand{\var}{{\em VAR}}
\newcommand{\meanVAR}{{\em meanVAR}}
\newcommand{\maxVAR}{{\em maxVAR}}
\newcommand{\minVAR}{{\em minVAR}}

\newcommand{\nqc}{{\em NQC}}
\newcommand{\qc}{{\em QC}}
\newcommand{\wig}{{\em WIG}}
\newcommand{\uWIG}{{\em U\_WIG}}
\newcommand{\smv}{{\em SMV}}
\newcommand{\uSMV}{{\em U\_SMV}}

\newcommand{\reference}{{\em REF}}
\newcommand{\bertpost}{{\em Bert-Ret}} 
\newcommand{\bertgen}{{\em Bert-Gen}-2A}
\newcommand{\bertgennorag}{{\em Bert-Gen}-1A$_{\text{no-rag}}$}
\newcommand{\bertgenrag}{{\em Bert-Gen}-1A$_{\text{rag}}$}
\newcommand{\bertgengeneral}{{\em Bert-Gen}-1A$_{(\cdot)}$}
\newcommand{\siamesereg}{{\em Internal Prober}}
\newcommand{\entropy}{{\em Uncertainty}}
\newcommand{\nli}{{\em Entailment}}

\newcommand{\nqopen}{{\em NQ-Open}}
\newcommand{\hotpot}{{\em HotpotQA}}
\newcommand{\trivia}{{\em TriviaQA}}

\newcommand{\accce}{{$\mathcal{Q}_{CE}$}}
\newcommand{\accefive}{{$\mathcal{Q}_{E5}$}}
\newcommand{\accnli}{{$\mathcal{Q}_{NLI}$}}
\newcommand{\accllm}{{$\mathcal{Q}_{LLM}$}}
\newcommand{\accem}{{$\mathcal{Q}_{EM}$}}
\newcommand{\gaince}{{\em Gain\textsubscript{$\mathcal{Q}_{CE}$}}}
\newcommand{\gainefive}{{\em Gain\textsubscript{$\mathcal{Q}_{E5}$}}}
\newcommand{\gaingeneral}{{\em Gain\textsubscript{Q()}}}

\newcommand{\truegain}{{\em True-Gain}}

\title{Predicting the Benefit of Retrieval Augmentation in Open-Domain Question Answering}

\author[1]{Or Dado}
\author[1,2]{David Carmel}
\author[1]{Oren Kurland}
\affil[1]{Technion, Haifa, Israel}
\affil[2]{Technology Innovation Institute (TII), Haifa, Israel}




\maketitle

\begin{abstract}
While retrieval-augmented generation has become a common approach for enhancing question answering systems, retrieval is not universally advantageous. We study the problem of predicting whether incorporating external retrieved information is likely to improve response quality for a given question. To this end, we evaluate a range of prediction methods 
that are based on retrieval signals, answer characteristics, and semantic consistency between generated responses and retrieved passages. 
We further devise a predictor that probes the LLM's internal state. Its prediction performance significantly narrows the performance gap between post-generation methods which are computationally demanding and pre-generation (post-retrieval) methods. We use the prediction methods to devise a selective retrieval framework that dynamically chooses between retrieval and non-retrieval generation modes per question. Experimental results demonstrate that selectively applying retrieval augmentation yields answer quality that transcends that of using retrieval for all queries.

\end{abstract}

\section{Introduction}

Retrieval-Augmented Generation (RAG) has become a widely adopted approach for enhancing the performance of Large Language Models (LLMs), particularly in question answering tasks \cite{lewis2020retrieval}. In this framework, content retrieved from an external corpus is incorporated into the prompt to ground the LLM’s responses in external knowledge and to mitigate hallucinations \cite{shuster-etal-2021-retrieval-augmentation}. Nevertheless, despite its overall effectiveness, RAG does not consistently yield benefits. In many cases, the retrieved content may be unhelpful or even distracting \cite{yoran2024making, yu2024chain}, potentially misleading the LLM and degrading answer quality \cite{amiraz-etal-2025-distracting}.

In this work, we study the following question: Can the benefit of applying RAG to a given question with respect to not applying it be predicted?
A reliable prediction of RAG gain would enable selective, instance-level application of RAG, thereby improving both system effectiveness and computational efficiency.

We begin by evaluating a range of pre-retrieval and post-retrieval predictors originally proposed for ad hoc retrieval tasks \cite{carmel2010estimating}, adapting them to the problem of RAG gain prediction. Our results show that pre-retrieval predictors are entirely ineffective in this setting, while certain post-retrieval predictors demonstrate limited yet non-negligible predictive capability. Furthermore, we show that supervised post-retrieval methods substantially outperform the unsupervised methods. Building on these findings, we propose a novel post-generation predictor that estimates RAG gain by incorporating the LLM-generated response in addition to pre-generation information. Specifically, 
the predictor models
semantic relationships among the question, retrieved passages, and generated answer. This predictor posts the highest prediction performance among all methods considered.

Post‑generation prediction approaches, while highly effective, incur substantial computational cost due to the need for explicit answer generation when estimating RAG gain. To address this limitation, we explore a novel gain predictor that ``probes'' the LLM’s internal state prior to answer generation in order to determine whether external knowledge is likely to be beneficial for a given question. Our empirical results show that this predictor substantially narrows the performance gap between pre-generation (post-retrieval) and post‑generation methods, thereby offering a practical and efficient alternative for post-generation gain prediction.

Additionally, we evaluate our predictors in a selective RAG setting, where the objective is to apply retrieval‑augmented generation on a per‑question basis only when an improvement in answer quality is predicted. 
Our experimental results demonstrate that selective RAG consistently outperforms the strategy of applying RAG uniformly to all questions.

In summary, our primary contribution is a systematic investigation of prediction strategies for estimating RAG gain in question answering;
the study includes novel predictors that post 
state-of-the-art performance. 
Through extensive experiments with two LLMs, two retrieval systems, and across three widely used question answering datasets, we demonstrate that supervised post-retrieval and post-generation predictors achieve strong performance and are promising candidates for integration into real-world RAG systems.

\section{Related Work}

The concept of RAG Gain was first introduced by Huly et al.~\cite{huly2025predicting} in the context of the text completion task.
Specifically, they formalized the relative performance improvement achieved by augmenting text with retrieved context before text completion. Huly et al. \cite{huly2025predicting} systematically evaluated a wide range of retrieval and generation-based signals as predictors of this gain.
Their analysis showed that post-retrieval and post-generation signals exhibit substantial predictive power. In this work, we adopt and modify their definition of RAG gain for open-domain question answering (Q\&A), a setting in which the interaction between retrieval relevance and answer quality is considerably complex.

Prior research on predicting RAG contribution for question answering has primarily focused on estimating retrieval utility for answer generation \cite{salemi2024evaluating, cuconasu2024power, tian2025relevance, tian2026predictingretrievalutilityanswer}. These studies indicate that retrieval utility largely depends on factors such as context length \cite{tian2025relevance}, the positioning of relevant passages within the retrieved context \cite{cuconasu2024power}, and potential conflicts between retrieved evidence and the LLM’s internal knowledge \cite{marjanovic-etal-2024-dynamicqa}. Dai et al.~\cite{daiseper} proposed predicting the RAG effect by estimating the LLM’s uncertainty during answer generation. 
Our \entropy{} predictor is based on the same observation, and is outperformed by our newly proposed predictors.

Lee at al. \cite{lee2025probing} explored alternative signals derived from an LLM’s internal knowledge. They trained a lightweight linear probe on the activations of selected attention heads during answer generation, empirically demonstrating that a well-trained model encodes signals correlated with the complexity of mathematical problems. Our novel RAG gain predictor, \siamesereg{}, also probes the LLM's internal state prior to answer generation, using a different approach, for predicting the relative
quality of an answer generated using RAG with respect to that generated without.

In a different direction, adaptive and selective RAG approaches 
have been proposed for deciding when retrieval should be invoked to improve answer quality. Methods such as Self-RAG~\cite{asai2024selfrag} and Active-RAG~\cite{jiang2023active} dynamically determine whether to retrieve additional context during generation, empirically demonstrating that retrieval is not universally beneficial. While these works underscore the importance of estimating RAG utility, they primarily focus on policy learning and generation control rather than on explicitly predicting RAG gain as a quantifiable outcome.
Jeong et al.~\cite{jeong-etal-2024-adaptive} proposed an adaptive question answering framework that predicts question complexity using a pre-retrieval classifier in order to dynamically choose the most suitable retrieval strategy for RAG. Similarly, Wang et al.~\cite{wang2025adaptive} studied retrieval decisions in conversational question answering, where the system determines at each dialogue turn whether retrieval augmentation should be applied.
Both approaches operate before retrieval and therefore cannot take into account retrieval quality or the LLM's internal knowledge. In contrast, some of our predictors leverage both factors to estimate the potential gain from applying RAG.

The work most related to ours is that of Tian et al.~\cite{tian2026predictingretrievalutilityanswer},
who also addressed the task of predicting the gain of applying RAG for QA. 
They studied several pre- and post-retrieval predictors, as well as a single post-generation predictor which amounts to our uncertainty-based predictor. We show that this predictor is substantially outperformed by our newly proposed post-generation predictors and by our predictor that probes the internal state of the LLM. We further show that post-generation predictors substantially outperform post-retrieval predictors which were the focus of Tian et al.'s ~\cite{tian2026predictingretrievalutilityanswer} work . While Tian et al.~\cite{tian2026predictingretrievalutilityanswer} used supervision to integrate existing predictors (either unsupervised or supervised predictors which target standard retrieval effectiveness), we use supervision to train novel methods to directly predict RAG gain. In addition, our predictor which probes the LLM state is the first of its kind, to the best of our knowledge, in ameliorating the (huge) performance gap between post-retrieval and post-generation prediction approaches.

\section{Task Definition}
\label{task_definition}
The challenge we address is the prediction of the RAG gain for question answering; that is, the gain of using retrieved content as context for the question at hand. We assume a classical RAG setting, in which the question is used to query an external corpus and the top scored retrieved results are concatenated with the question to prompt the LLM for answer generation.

Following Huly et al. \cite{huly2025predicting}, we define the relative (log) gain of using RAG with a retrieved context $C$, for answering question $q$, as:
\begin{equation}
    \label{eq:gain}
    Gain(C |q, r, \mathcal{Q}, \theta) = \log\frac{\mathcal{Q}(\theta(q;C),r)+\epsilon}{\mathcal{Q}(\theta(q),r) +\epsilon},
\end{equation}
where $\theta(x)$ is the LLM's response to the prompt $x$; $q;C$ is the concatenation of $q$ with $C$, $\mathcal{Q}(a,r)$ is the quality metric used to measure the quality of generated answer $a$ with respect to the reference (ground truth)
answer $r$, and $\epsilon$ ($\epsilon=1\times 10^{-12}$ in our implementation) supports numerical stability. The task we address below is predicting the gain defined in Equation \ref{eq:gain} with no knowledge of the reference answer $r$.

\subsection{Quality Metrics}
\label{subsec: quality-metrics}
Most factoid question answering benchmarks, including Natural Questions (NQ) \cite{kwiatkowski2019natural}, TriviaQA \cite{joshi2017triviaqa}, HotpoQA \cite{yang2018hotpotqa}, and PopQA \cite{mallen2023popqa}, rely on Exact Match (EM) and token-level F1 as their primary evaluation metrics. EM is a binary, factoid-level accuracy measure that evaluates whether the reference answer exactly matches, or is fully contained, within the generated answer. F1 quantifies surface-level token overlap between the generated and reference answers. While these metrics are simple and widely adopted, they fail to capture important semantic aspects of answer quality. In particular, they do not fully account for the presence of redundant information or misinformation. 
Moreover, they do not assess whether the generated response sufficiently satisfies the underlying information need. As a result, EM and F1 are limited in their ability to accurately evaluate the quality of LLM-generated answers and, by extension, the true gain achieved by RAG \cite{es2024ragas}.

To address these limitations, a common approach for evaluating answer quality is the LLM-as-a-judge paradigm \cite{gu2024survey}, in which typically a strong LLM is tasked with assessing answer quality by comparing the generated response with a reference answer. Although this approach has been shown to correlate highly with human judgments, it is sensitive to the choice of the judging LLM, as such models may introduce their own biases \cite{zheng2023judging}. Moreover, LLM-based evaluation incurs substantial computational cost, limiting its scalability.

The high cost of LLM-as-a-judge approaches motivates the need for alternative metrics that approximate the semantic acuity of powerful LLM evaluators while avoiding their computational overhead. 
Prior work \cite{balamurali2025revisiting} demonstrated that answer quality can be effectively estimated using BERT‑based shallow models, achieving high correlation with human judgments and performance comparable to GPT‑4o\footnote{\url{https://chat.chatbot.app/gpt4o}}, a state‑of‑the‑art LLM, when employed as a judge.

\begin{figure}[t] 
\includegraphics[width=0.89\textwidth]{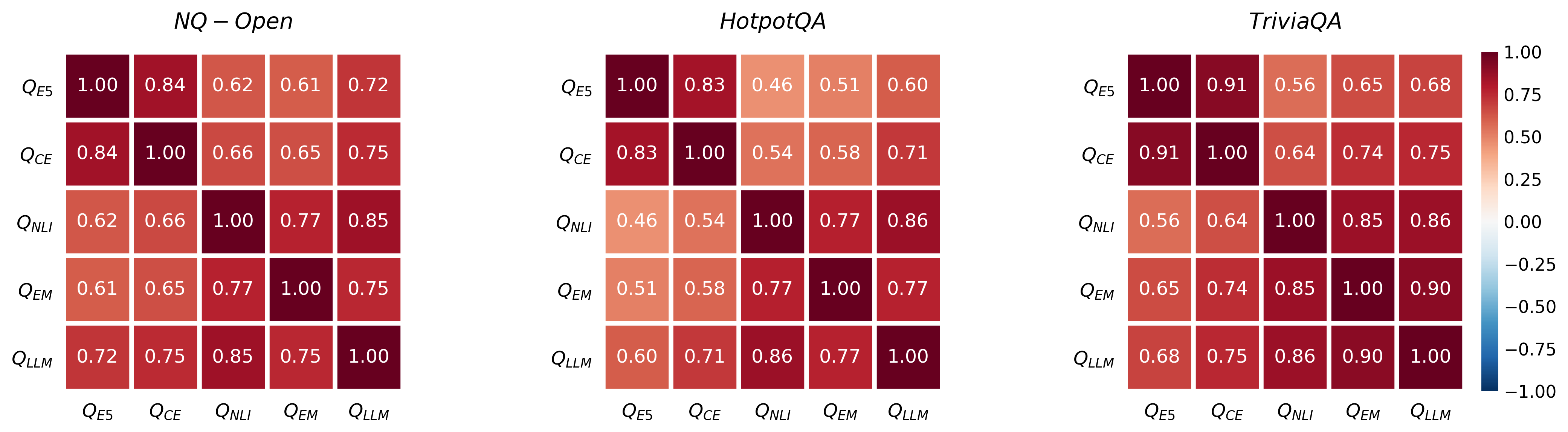} \\
\includegraphics[width=0.89\textwidth]{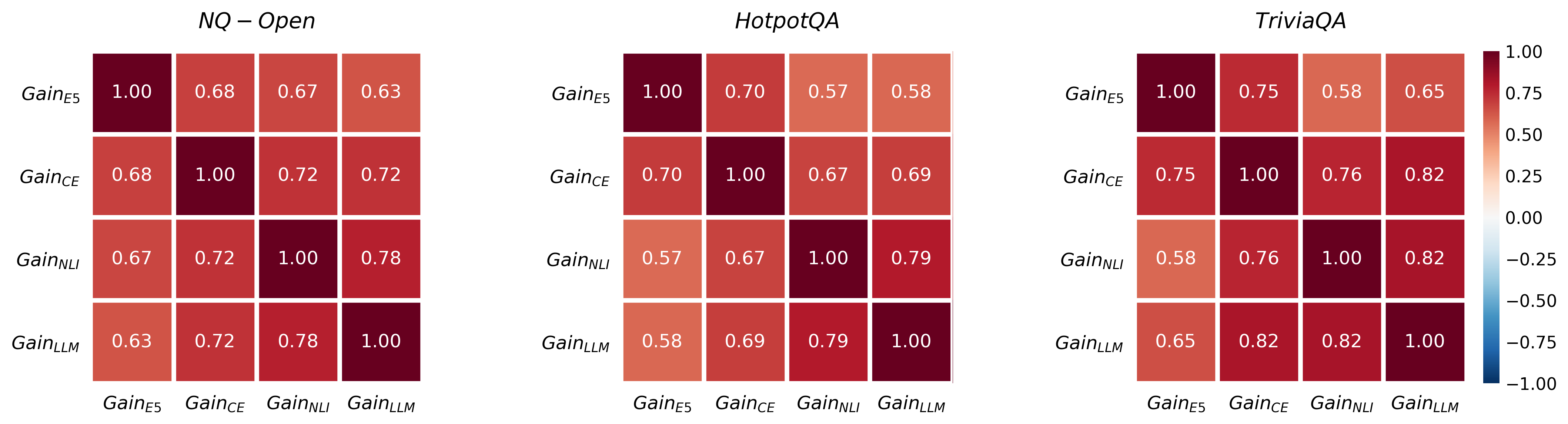} 
\caption{\label{fig:quality-heatmap}\small
(Top): Pairwise Pearson correlations between the quality metrics across \nqopen{}, \hotpot{}, \trivia{}. 
(Bottom): Pairwise Pearson correlations between the RAG gains estimated using the quality metrics. The gain is measured between the {\em with-rag} and {\em with-no-rag} answers generated by Falcon3, accompanied with BM25 retrieval over Wikipedia. 
}
\end{figure}

In this work, we experiment with three shallow answer quality metrics that assess semantic alignment between the generated answer and the reference answer:

\begin{itemize}[leftmargin=0pt] 

\item \accefive{}: The cosine similarity between embedding vectors of the generated answer and the reference answer, where embeddings are computed using the pre-trained E5-large-v2 model \cite{wang2022text}.

\item \accce{}:
A semantic association score between the generated and reference answers computed using a pre-trained cross-encoder based on RoBERTa-large\footnote{\url{https://huggingface.co/sentence-transformers/stsb-roberta-large}} \cite{liu2019roberta}. The raw cross-encoder score is normalized to the range $[0,1]$ using a sigmoid function.

\item \accnli{}:
Answer quality is framed as a natural language inference (NLI) problem; specifically, whether the generated answer is semantically entailed by the reference answer \cite{balamurali2025revisiting}. We employ the \texttt{DeBERTa-v3-large-NLI} model\footnote{\url{MoritzLaurer/DeBERTa-v3-large-mnli-fever-anli-ling-wanli}} \cite{laurer2023building}, which is trained on a large and diverse collection of NLI datasets, to estimate question quality based on entailment probability.
\end{itemize}
Figure~\ref{fig:quality-heatmap} (Top) presents pairwise Pearson correlation coefficients between the quality metrics, measured over test sets of 3,600 Q\&As each, sampled from the validation sets 
of  three question answering datasets: \nqopen{}, \hotpot{}, \trivia{}. The metrics measure the quality of answers generated by Falcon3-10B-Instruct~\cite{falcon2024falcon},
with respect to the reference (ground-truth) answers. 
For reference, we also correlate the quality metrics with \accem{} 
(exact match measured via answer inclusion)
and with \accllm{}, an LLM-as-a-judge metric implemented through the DeepEval framework\footnote{\url{https://deepeval.com/}};
we use GPT-4o as a judge. (The prompt used for answer judgment is provided  in Appendix \ref{app:prompts}.)

The strong correlation observed in Figure~\ref{fig:quality-heatmap} (Top) 
between all quality metrics indicates a high degree of assessment
agreement. However, the variance of correlation values (e.g., ranging from $0.61$ to $0.84$) suggests that the metrics are not fully interchangeable and capture complementary aspects of quality. 
Interestingly, the metrics are split into two groups across all datasets; \accefive{} and \accce{} are highly correlated as both measure semantic similarity, while \accnli{} is highly correlated with \accem{} and \accllm{}.

Figure \ref{fig:quality-heatmap} (Bottom) illustrates the pairwise Pearson correlation between the RAG gains while measured using the different quality metrics. We use Falcon3-10B-Instruct, 
with and without RAG, as our  answer generator over the three test sets; BM25~\cite{robertson1992okapi} is used for retrieval over Wikipedia dump \cite{wikipedia2018}. (The RAG implementation is further detailed in Section \ref{seq:eperiments}.).
As indicated by the strong correlations between the gain estimates,
there is substantial agreement across methods regarding the potential benefit of applying RAG for individual questions.

\subsection{RAG-gain Analysis}
\begin{figure}[htbp]
    \centering
    \includegraphics[width=0.48\textwidth]{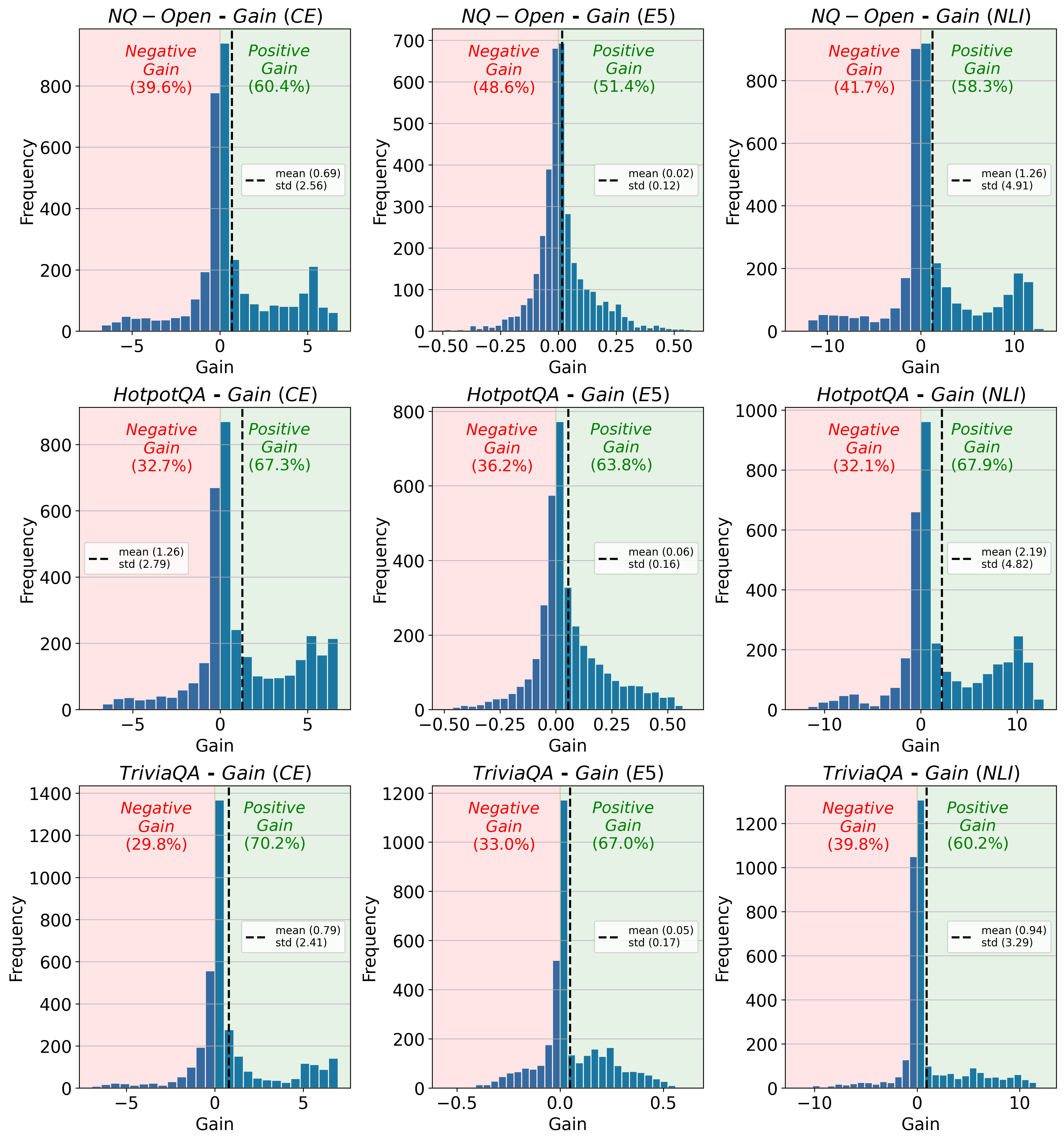}
    \caption{\small
    RAG gain distribution across the three Q\&A  test sets, each comprising 3,600 Q\&A pairs sampled from \nqopen{}, \hotpot{}, and \trivia{}. The backbone LLM is Falcon-3-10B, augmented with RAG based on BM25 retrieval over a Wikipedia dump. 
}
    \label{fig:gain_distributions}
\end{figure}

RAG benefits question answering by improving answer quality, grounding responses with external knowledge, and reducing hallucinations. In this work, we focus specifically on improvements in answer quality resulting from RAG usage, as quantified by our gain metric (Equation~\ref{eq:gain}). 
Figure~\ref{fig:gain_distributions} presents the distribution of RAG gain based on the three quality metrics defined above and across the test sets of the three Q\&A datasets.

We can see in Figure ~\ref{fig:gain_distributions} that across all datasets and quality metrics, the average RAG gain is positive, highlighting the overall effectiveness of RAG for question answering. 
While the magnitude of the average gain varies by dataset, the gain distributions exhibit a consistent pattern: a pronounced central peak near zero accompanied by heavy tails skewed toward positive values. This indicates that for a significant fraction of the questions, RAG yields little or no improvement in answer quality (e.g. >30\% of the questions have negative gain in HotpotQA).
These findings suggest that indiscriminately applying RAG to all questions may introduce unnecessary latency and computational overhead. Consequently, they motivate the task of RAG gain prediction --- by estimating the expected gain in advance, a system can adopt a selective RAG strategy, invoking retrieval only when it is likely to be beneficial. 

\section{Gain Prediction Methods}
In what follows we survey a few prediction methods which were originally proposed for ad hoc retrieval. In addition, we propose a few novel RAG gain predictors for question answering.
We study three types of prediction methods.
Pre-retrieval predictors analyze only the question and
corpus-level statistics. Post-retrieval predictors additionally analyze the retrieved passages. Post-generation predictors analyze the question, the passages, and the generated answer.
Since the question serves as the query used for retrieval in RAG, we will use throughout the rest of this section the terms ``query'' and ``question'' interchangeably.

\subsection{Pre-Retrieval Predictors}
Pre-retrieval predictors are unsupervised measures used in ad hoc
retrieval to estimate retrieval effectiveness before retrieval is performed \cite{hauff2008survey}. These methods estimate the relationship between query
terms and their statistical properties in the document corpus. 
\begin{itemize}[leftmargin=0cm]
    
    \item \textbf{\scq{}:} The Similarity to Collection  Query (SCQ) method estimates how well a query will perform based on its similarity to the 
    collection. It measures the sum of \emph{tf-idf}  weights of query terms \cite{zhao2008effective}. 
    
    \item \textbf{\idf{}:} This is the average 
    idf values 
    of the terms in a query \cite{kwok1996new}. It is based on the premise that rare or highly specific terms are more effective at distinguishing relevant documents from the rest of the collection.
    
    \item \textbf{\var{}:} This is the average of the variances of the
     query term's tf-idf weights across the documents in which 
     they appear \cite{zhao2008effective}. A higher variance suggests that a term has stronger discriminative power, making it a better indicator of effective retrieval. 
\end{itemize}

\subsection{Post-Retrieval Predictors}
In ad hoc retrieval, post-retrieval predictors analyze the retrieved
list, in addition to the query and corpus statistics \cite{carmel2010estimating}. Here we study
whether the predicted list effectiveness is correlated with the gain
attained by using the highest ranked passages for augmentation. We
explore the following 
post-retrieval predictors:

\subsubsection{\textbf{Unsupervised predictors}}
    \begin{itemize}[leftmargin=0cm]
        \item{\textbf{\wig{}:}} The difference between the average retrieval score of the top-$k$ passages in the retrieved list, $L$, and the corpus retrieval score; $k$ is a parameter \cite{zhou2007query}. The assumption is that the higher the difference, the more effective the retrieval as the corpus is essentially a pseudo non-relevant document. We also study a variant, \uWIG{}, which uses the average of top scores without the corpus score regularization.
        \item{\textbf{\nqc{}:}} The standard deviation of retrieval scores of the top-$k$ passages in $L$ normalized by the corpus retrieval score \cite{shtok2012predicting}. We also study a variant, \qc{}, which does not use the corpus normalization. The assumption is that high standard deviation indicates reduced query drift and, hence, improved retrieval effectiveness.
        \item{\textbf{\smv{}:}} Combines the mean and variance of the retrieval scores of the top-$k$ passages in $L$; the corpus retrieval score is used for normalization \cite{tao2014query}. We also study a variant, \uSMV{}, which does not apply corpus normalization.
        \item{\textbf{\reference{}:}} 
        The difference between the retrieved list and a  high-quality reference list \cite{shtok2016query}. 
        The assumption behind this predictor is that high similarity to the reference list indicates higher quality hence a higher RAG gain. 
        Given the raw search results to be used for RAG, we re-rank the top-100 retrieved passages with a pre-trained reranker (BAAI/bge-reranker-v2-m3 reranker\footnote{\url{https://huggingface.co/BAAI/bge-reranker-v2-m3}} \cite{chen2024bge}) to be used as a high-quality reference list. We then use the Rank Biased Overlap (RBO) \cite{webber2010similarity} between the original list and the re-ranked list as the prediction score.
    \end{itemize}

\subsubsection{\textbf{Supervised predictors}}

    \begin{itemize}[leftmargin=0cm]
        
        \item{\textbf{\bertpost{}:}} A supervised BERT-based predictor, inspired by the BERT-QPP model \cite{arabzadeh2021bertqpp}, which  models the textual association of the query with top-retrieved passages for RAG gain prediction. We provide the query $q$ and the top passages  as input to ModernBERT-large\footnote{\url{https://huggingface.co/answerdotai/ModernBERT-large}} \cite{warner2024smarter}, a highly effective variant of BERT
          with a long window context. A linear regression layer is added on top of the BERT encoder, and the entire model is trained end to end using a Mean Squared Error (MSE) loss to predict the RAG gain. The training process is further detailed in Section \ref{seq:eperiments}.
        
    \end{itemize}

\subsection{Post-Generation Predictors}
The predictors discussed so far operate before the LLM is prompted to generate an answer. We next leverage the LLM's output, with and without RAG, to devise post-generation predictors.

\subsubsection{\textbf{Unsupervised predictors}}

\begin{itemize}[leftmargin=0cm]
    \item{\textbf{\entropy{}:}}     
    This predictor follows the assumption that high reduction in answer generation uncertainty reflects high quality gain. For each generated sequence (i.e., answer), we compute token-level entropy from the LLM's probability distribution output, taking the maximum entropy across the sequence as a  measure of model uncertainty \cite{kuhn2023semantic}. Let $A = (t_1,\ldots,t_n)$ be the sequence of tokens generated by the LLM, and $\mathcal{H}(t)$  be the entropy of the token distribution induced by the LLM while token $t\in A$ is generated. We measure the reduction in uncertainty between no-RAG and RAG generated answers, to be used for gain prediction:
    \[
    \displaystyle{
    \max_{t\in A_{\text{no-rag}}}{\mathcal{H}(t)} - \max_{t\in A_{\text{rag}}}{\mathcal{H}(t)}};
    \]
$A_{\text{no-rag}}$ and $A_{\text{rag}}$ are the answers generated by the LLM, without and with RAG, respectively.    

\item{\textbf{\nli{}:}}   
This predictor is based on the assumption that higher faithfulness of the generated answer to the RAG context leads to better quality. 
We adopt an NLI model based on ModernBERT-large\footnote{\url{https://huggingface.co/MoritzLaurer/ModernBERT-large-zeroshot-v2.0}}, 
which is pre-trained on Natural Language Inference tasks and supports longer input contexts. 
For each question, the top-$5$ retrieved passages are concatenated to the question to form the premise: \emph{Premise} $= (q;p_1, \dots, p_5)$,  while the hypothesis is the generated answer concatenated with the question: \emph{Hypothesis} $= (q;A)$.
We compute the entailment score twice: once for the RAG-generated answer, $Ent(A_{\text{rag}})$, and once for the no-RAG answer, $ Ent(A_{\text{no-rag}})$. The predicted gain is then defined as the difference between the two entailment scores:
\[
\displaystyle{Ent(A_{\text{rag}}) - Ent(A_{\text{no-rag}})}.
\]
      
 \end{itemize}
 
 \subsubsection{\textbf{Supervised predictors}}     
\label{post_gen_supervised}
\begin{itemize}[leftmargin=0cm]
\item{\textbf{\em Bert-Gen}:}
The model 
has the same architecture as \bertpost{}, but is additionally trained using answers generated by the LLM, both with and without RAG.

We experiment with three variants of this predictor. The \bertgennorag{} variant incorporates the answer generated without RAG, together with the question and retrieved passages. The \bertgenrag{} variant incorporates the answer generated with RAG, while \bertgen{} incorporates both the RAG and no‑RAG answers for gain prediction. The key distinction among these variants lies in their operational assumptions. The variants that incorporate only a single generated answer can make an early decision regarding the necessity of generating the alternative response. Specifically, \bertgennorag{} can be used to identify cases in which the initial answer is of sufficient quality, allowing RAG‑based answering to be skipped.  In contrast, \bertgen{} explicitly compares the two answers and can therefore make a more informed decision about which is preferable, 
with the cost of two answer generations.
\end{itemize}

\subsection{Semi-Generation (\siamesereg{})}
\label{sec:siamese_predictor}

We introduce \siamesereg{}, a novel gain predictor that directly estimates RAG gain from the internal representations of the LLM, without requiring full auto-regressive generation. By inspecting the final hidden states of the LLM during the forward pass of the prompt, we effectively probe the model’s internal representation space, enabling estimation of the merit of using the retrieved context with respect to answer quality, without generating any output tokens.

\begin{figure}[t]
    \centering
    \includegraphics[width=0.895\textwidth]{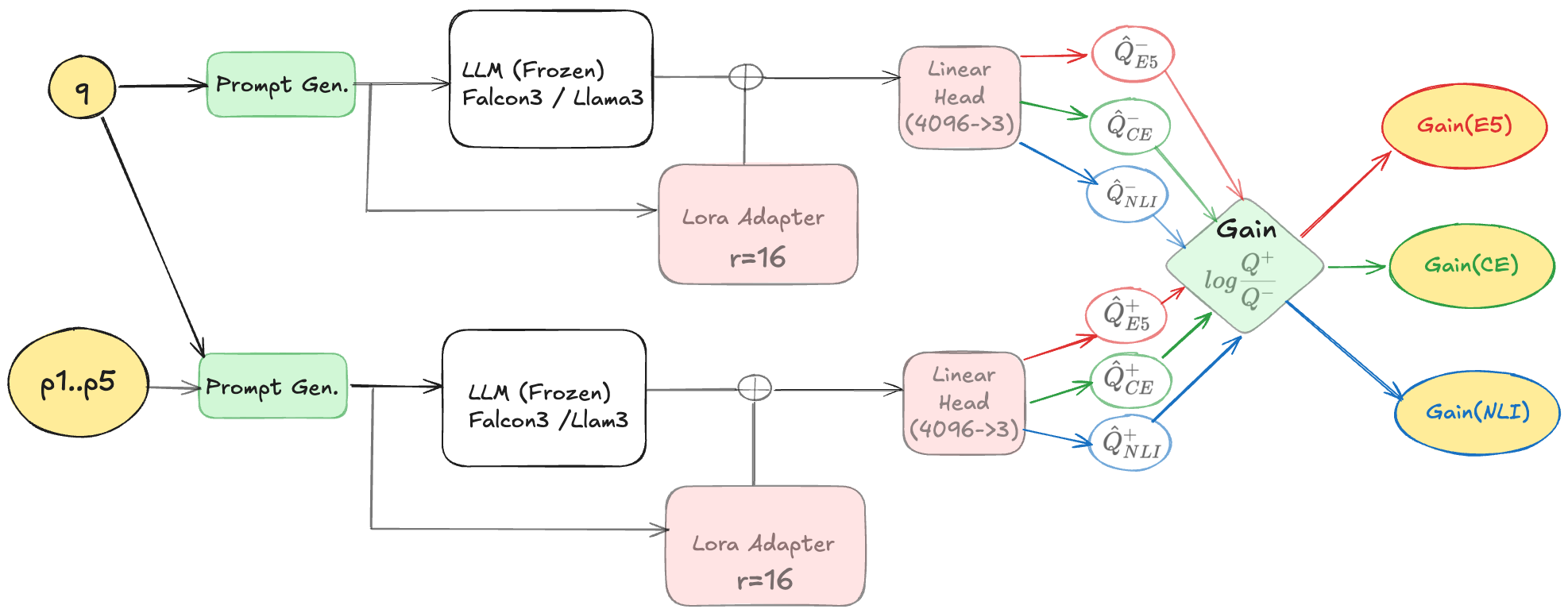}
    \caption{\small
    \siamesereg{} architecture for RAG gain prediction. Two independent Lora adapters, 
    sharing a frozen LLM backbone, are each fine-tuned with separate LoRA weights. The upper branch processes the query $q$ alone, producing $\hat{Q}^-_{E5}$, $\hat{Q}^-_{CE}$, $\hat{Q}^-_{NLI}$; the lower branch processes $q$ together with retrieved passages, producing $\hat{Q}^+_{E5}$, $\hat{Q}^+_{CE}$, $\hat{Q}^+_{NLI}$. Gain, a log-ratio combiner, computes $\log \frac{Q^+}{Q^-}$ per metric, yielding three RAG gain predictions $Gain(E5)$, $Gain(CE)$, and $Gain(NLI)$.
    }
    \label{fig:internal_state_prober_architecture}
\end{figure}

Figure~\ref{fig:internal_state_prober_architecture} presents the architecture of \siamesereg{}. The model is explicitly designed to capture the marginal gain of retrieval through a dual structure. It operates over a single frozen LLM with two parallel input streams: the question augmented with retrieved passages ($q ; p_1 \ldots p_5$), and the question alone ($q$). To avoid duplicating the full model, we train two independent Low-Rank Adaptation (LoRA) modules \cite{hu2022lora}, one per branch. This parameter-efficient design allows each branch to specialize to its respective input distribution, while sharing the underlying model parameters.

Each branch performs a single forward pass, from which we extract the final hidden state representation. These representations are passed through linear regression heads, producing three scalar outputs that can be interpreted as predicted ``quality'' scores corresponding to the quality functions defined in Section \ref{subsec: quality-metrics}.
The three scores from both branches are are then respectively paired to approximate the  expected gains using Equation~\ref{eq:gain}.
The model is trained using MSE loss with respect to the ground-truth gains.
Further details on the model training are given in Section \ref{sec:training}.

Typical internal probing methods train only a lightweight regression layer on top of a frozen LLM (e.g., \cite{lee2025probing}), so that task learning is confined to the probe while the LLM’s internal representations remain unchanged. In contrast, \siamesereg{} actively adapts the LLM by training LoRA adapters, allowing the model’s internal hidden states to be reshaped in accordance with the target task.
As a result, the learned representations explicitly encode task‑specific parametric knowledge relevant to RAG gain prediction, rather than merely exposing signals that happen to be present in a fixed representation.

At inference time, \siamesereg{}  replaces the standard auto-regressive 
generation process which require $|A_{\text{no-rag}}| +|A_{\text{rag}}|$ sequential forward passes by \bertgen{}, with only two parallel forward passes. As a result, \siamesereg{} predicts RAG gain while significantly reducing computational cost, making it well-suited for large-scale and latency-sensitive question answering pipelines.

\section{Experimental Setup}
\label{seq:eperiments}

Our experimental setup follows standard practice in Query Performance Prediction (QPP) for ad-hoc retrieval \cite{carmel2010estimating}, adapted here to the task of
RAG gain prediction for question answering.
Prediction quality is measured by the correlation between actual gain and predicted gain. 

\paragraph{{\bf LLMs.}}
We experiment with two mid-sized, instruction-tuned LLMs for question answering: Falcon-3-10B-Instruct\footnote{\url{https://huggingface.co/tiiuae/Falcon3-10B-Instruct}} \cite{falcon2024falcon} and Llama-3.1-8B-Instruct\footnote{\url{https://huggingface.co/meta-llama/Llama-3.1-8B-Instruct}} \cite{dubey2024llama}.

\paragraph{{\bf Retrieval.}}
Retrieval is performed over a Wikipedia dump corpus (Dec. 20, 2018) \cite{wikipedia2018}, where articles are segmented into disjoint 100-word passages. We evaluate two distinct retrieval paradigms: sparse retrieval and dense retrieval. For sparse retrieval we use Okapi BM25 \cite{robertson1992okapi} via the Pyserini implementation \cite{lin2021pyserini} with default parameter values. For dense retrieval we employ the E5-large-v2 model\footnote{\url{https://huggingface.co/intfloat/e5-large-v2}} for text embedding, indexing the same Wikipedia dump via the Faiss similarity search library \cite{johnson2019billion}.
For every question, we retrieve $k=100$ passages. 
For RAG, the top-5 passages are used to augment the question for answer generation. 

\paragraph{{\bf Datasets.}}
We conduct experiments over three open-domain Question Answering (QA) datasets: \texttt{\nqopen{}} \cite{kwiatkowski2019natural}, \texttt{\hotpot{}} \cite{yang2018hotpotqa}, and \texttt{\trivia{}} \cite{joshi2017triviaqa}. 
All these datasets provide a large set of factual questions associated with reference answers, considered as ground-truth answers.
To address inconsistencies where original datasets lack test splits or provide incomplete reference answers for some of the test questions, we enforce a standardized partitioning strategy. From the train split of each dataset, we randomly sample $30,000$ questions and divide them into an $80-20$ train/validation split. 
The training set ($24,000$ questions) is used to train the supervised predictors. The validation set ($6,000$ questions) is used for hyperparameter tuning.  
For the test sets, we randomly sample $3,600$ questions from the validation split of each dataset, ensuring complete separation between training and testing data.

\paragraph{{\bf Evaluation:}}
To be consistent with established conventions in QPP literature, we measure prediction quality using the Pearson correlation ($\rho$) between the predicted gain and the actual gain \cite{carmel2010estimating}. We found the relative prediction quality patterns to be consistent with those attained by using Spearman correlation or Kendall's tau. The Spearman and Kendall's tau numbers are omitted to avoid clusttering the presentation and as they convey no additional insight. To determine the statistical significance of correlation differences, 
we employ Williams' two-tailed t-test~\cite{williams1959} ($p = 0.05$), which accounts for the fact that two correlations computed on the same sample against the same target are statistically dependent. We apply
Bonferroni correction for multiple hypothesis testing \cite{miller1981normal}.
We report prediction performance for RAG gain, as formulated in Equation \ref{eq:gain}, based on the three answer-quality metrics: 
\accefive{}, \accce{} and \accnli{} . 

\subsection{Training}
\label{sec:training}
To train the supervised models, RAG gain prediction is formulated as a supervised regression problem. 

\subsubsection{{\bf Cross-Encoder-based Predictors:}}
The cross-encoder-based predictors (\bertpost{}, \bertgengeneral{} and \bertgen{}) use ModernBERT-large as the backbone encoder which enables joint attention across the question, retrieved passages, and generated answers.
For \bertpost{}, the input sequence consists of the question concatenated with the top-$5$ retrieved passages: 
            $([CLS]\ q\ [SEP]\ p_1\ [SEP]\dots\ \ p_5).$
For \bertgengeneral{}, we additionally append one of the 
generated answers ($A_{\text{no-rag}}~\text{or}~ A_{\text{rag}}$):
        $
            ([CLS]\ q\ [SEP]\ A_{(\cdot)}\ [SEP] p_1\ [SEP]\dots\ \ p_5).
        $
For \bertgen{}, we append the two generated answers:
        $$
        ([CLS]\ q\ [SEP]\ A_{\text{no-rag}}\ [SEP]\ A_{\text{rag}}\ [SEP]\ p_1\ [SEP] \dots\ \ p_5).
        $$
ModernBERT-large produces a $1,024$-dimensional contextual embedding vector for each token. We use the \texttt{[CLS]} token representation as a pooled representation of the entire sequence and feed it into a linear regression layer that outputs a single scalar corresponding to the predicted \textit{RAG gain}.
The model is fine-tuned end-to-end over the training data, using MSE loss between the predicted gain and the actual gain. Training is performed for two epochs using the AdamW optimizer \cite{loshchilov2018decoupled} with a learning rate of $5 \times 10^{-5}$ and a batch size of $16$. The maximum input length is set to $8,192$ tokens, which comfortably accommodates the question, 5 retrieved passages, and 2 generated answers.
We train a separate model for each combination of LLM, retriever, dataset, and quality metric, resulting in $2\cdot 2 \cdot 3 \cdot 3 = 36$ distinct configurations.

\subsubsection{{\bf \siamesereg{}:}}
As introduced in Section~\ref{sec:siamese_predictor}, this predictor leverages the LLM (Falcon3 or Llama3) as a shared, frozen backbone, with two trainable parameter-efficient Lora adapters.
Each training instance is a tuple containing the question, the top-5 retrieved passages, and the reference answer used for calculating the target RAG gain.
The two inputs, ($[q]$ and $[q;p_1..p_5]$), are processed independently through the two LoRA adapters attached to the same backbone LLM, and the two distinct linear heads. Lora is applied to 
the transformer's query, value, output, and gate projection matrices ($W_q, W_v, W_o$, $W_{\text{gate}}$),
with rank $r=16$,  a scaling factor $\alpha=32$, 
and dropout $0.1$ (See Section \ref{subsec:hyper} on hyper-parameters tuning). The three linear head outputs are coupled at the final stage (after Softplus transformation), to compute the predicted gain via log-difference. The model is trained using MSE with respect to the target gain.

We jointly predict gains for all three quality metrics in parallel, which serves as a form of multi-task regularization and improves robustness. Since each quality metric operates on a different scale, the corresponding (true) gains also differ in magnitude, which can lead to imbalance in the loss. 
Hence, we normalize each gain by its mean and standard deviation prior to training.
    
Training is performed for 2 epochs with a batch size of $16$, using AdamW with a 
learning rate of $5 \times 10^{-5}$ 
and weight decay of $0.01$. To support long input sequences and hardware constraints, we employ BFloat16 precision
for memory reduction.

\subsection{Hyper-Parameters Tuning}
\label{subsec:hyper}
Hyper-parameter values were selected, for each configuration, by maximizing the total correlation of predicted gain with the actual gain across the validation set. The total correlation is defined as the sum of gain correlations computed separately for each of the three quality measures.

Pre-retrieval predictors do not involve any hyperparameter tuning. 
For post-retrieval predictors (\wig{}, \uWIG{}, \nqc{}, \qc{}, \smv{}, and \uSMV{}), the length of the 
retrieved result list is selected from $\{1, 2, 3, 4, 5, 10, 20, 30, 40, 50\}$. 
For the \reference{} predictor which relies on RBO, the depth parameter $L$ which controls the prefix length of the lists to be compared is selected from $\{1, 5, 10, 20, \dots, 100\}$, and the decay rate parameter $p$ is selected from $\{0.9, 0.95, 0.99\}$. 

For 
post-retrieval and post-generation predictors,
we experiment with the number of top passages to be used for augmentation, selected from $\{1, 2, 3, 4, 5\}$. Results consistently indicate an advantage for larger values; accordingly, we set the number of top passages for augmentation to be 5.

\entropy{} operates solely on the LLM's output probability distribution, producing a token-level uncertainty signal for all tokens in the generated answer. To aggregate these signals into a single sequence-level score, we evaluate five pooling strategies over the token entropy values $\{\mathcal{H}(t)\}_{t \in Ans}$: arithmetic mean, geometric mean, harmonic mean, min, and max. Max pooling yields the strongest predictive signal across all evaluated configurations under the hyper-parameter selection criterion.

For \siamesereg{}, we fine-tune the main LoRA hyper-parameters.
We experiment with learning rate values in $\{1 \times 10^{-5}, 5 \times 10^{-5}, 1 \times 10^{-4}\}$, LoRA rank values $r\in \{8, 16, 32\}$, and the scaling factor $\alpha\in\{16, 32, 64\}$. In addition, we evaluate different adapter placement strategies, comparing injection across all transformer layers, or restricting adaptation to final (final-12 or final-20) layers only. The best configuration found is to apply LoRA adapters to all layers with $r=16$, $\alpha=32$, and a learning rate of $5 \times 10^{-5}$.

\section{Results}
Table~\ref{tab:combined_results} summarizes the gain prediction quality of all predictors across the different configurations, as measured by Pearson correlation with the true gain.

\begin{table*}
\centering
\footnotesize
\caption{\small Pearson correlation between RAG gain predictions and actual gains measured using the three quality metrics (\accefive{}, \accce{}, \accnli{}), across the \nqopen{}, \hotpot{}, and \trivia{} test sets. The answers were generated by the two LLMs (\textbf{Falcon3}, \textbf{Llama-3.1}) using two retrieval methods (BM25 and E5) for RAG. 
$\dagger$ indicates statistically significant better prediction than all predictions appearing above in the corresponding column,
per retrieval method. 
Boldface marks the best performance per column for a retrieval method.
}
\label{tab:combined_results}
\setlength{\tabcolsep}{3pt}
\scriptsize
\begin{tabular}{l||ccc|ccc||ccc|ccc||ccc|ccc}
\toprule
& \multicolumn{6}{c||}{\textbf{\nqopen{}}} & \multicolumn{6}{c||}{\textbf{\hotpot{}}} & \multicolumn{6}{c}{\textbf{\trivia{}}} \\
\midrule
&  \multicolumn{3}{c|}{\textbf{Falcon}} & \multicolumn{3}{c||}{\textbf{Llama}} & \multicolumn{3}{c|}{\textbf{Falcon}} & \multicolumn{3}{c||}{\textbf{Llama}} & \multicolumn{3}{c|}{\textbf{Falcon}} & \multicolumn{3}{c}{\textbf{Llama}} \\
Predictor & \accefive{} & \accce{} & \accnli{} & \accefive{} & \accce{} & \accnli{} & \accefive{} & \accce{} & \accnli{} & \accefive{} & \accce{} & \accnli{} & \accefive{} & \accce{} & \accnli{} & \accefive{} & \accce{} & \accnli{} \\
\midrule

\multicolumn{19}{c}{\color{gray}\textbf{BM25 Retrieval}} \\
\midrule
\idf{} & .141 & .121 & .120 & .109 & .084 & .103 & -.013 & .030 & .009 & .018 & .049 & .033 & -.045 & -.029 & -.000 & .042 & .063 & .073 \\
\scq{} & .120 & .110 & .108 & .093 & .082 & .090 & .019 & .036 & .036 & .073 & .051 & .053 & -.019 & -.009 & .022 & .033 & .052 & .066 \\
\var{} & .125 & .108 & .118 & .098 & .071 & .080 & -.007 & .030 & .016 & .015 & .053 & .015 & -.023 & -.011 & .013 & .046 & .061 & .069 \\

\midrule
\nqc{} & .108 & .081 & .083 & .098 & .067 & .085 & .067 & .107 & .107 & .080 & .063 & .054 & .023 & .013 & .024 & .074 & .066 & .087 \\
\qc{} & .199 & .185 & .190 & .154 & .117 & .147 & .112 & .106 & .136 & .120 & .083 & .091 & .151 & .120 & .103 & .132 & .115 & .122 \\
\wig{} & .158 & .149 & .136 & .157 & .100 & .122 & .089 & .095 & .124 & .180 & .117 & .116 & .061 & .036 & .063 & .144 & .107 & .115 \\
\uWIG{} & .141 & .136 & .131 & .164 & .095 & .117 & .122 & .107 & .143 & .190 & .121 & .135 & .103 & .071 & .070 & .127 & .101 & .105 \\
\smv{} & .108 & .081 & .083 & .100 & .067 & .087 & .067 & .107 & .107 & .081 & .063 & .054 & .022 & .013 & .024 & .053 & .042 & .065 \\
\uSMV{} & .199 & .183 & .189 & .151 & .116 & .146 & .091 & .081 & .108 & .120 & .083 & .091 & .150 & .118 & .099 & .130 & .114 & .122 \\
\reference{} & .255 & .219 & .253 & .305 & .190 & .226 & .173 & .176 & .213 & .255 & .146 & .203 & .214 & .205 & .190 & .286 & .226 & .236 \\
\bertpost{} & .450$^\dagger$ & .285$^\dagger$ & .428$^\dagger$ & .447$^\dagger$ & .292$^\dagger$ & .312$^\dagger$ & .375$^\dagger$ & .324$^\dagger$ & .405$^\dagger$ & .484$^\dagger$ & .288$^\dagger$ & .327$^\dagger$ & .304$^\dagger$ & .425$^\dagger$ & .403$^\dagger$ & .121 & .319$^\dagger$ & .330$^\dagger$ \\

\siamesereg{} & .530$^\dagger$ & .446$^\dagger$ & .484$^\dagger$ & .603$^\dagger$ & .372$^\dagger$ & .416$^\dagger$ & .484$^\dagger$ & .408$^\dagger$ & .476$^\dagger$ & .578$^\dagger$ & .396$^\dagger$ & .461$^\dagger$ & .547$^\dagger$ & .532$^\dagger$ & .517$^\dagger$ & .546$^\dagger$ & .451$^\dagger$ & .469$^\dagger$\\

\midrule
\entropy{} & .419 & .251 & .330 & .354 & -.022 & .168 & .371 & .212 & .280 & .419 & .006 & .196 & .545 & .428 & .388 & .547 & .283 & .296 \\

\nli{} & .214 & .186 & .224 & .154 & .077 & .100 & .276 & .241 & .335 & .320 & .165 & .274 & .366 & .412 & .408 & .317 & .250 & .284 \\

\bertgennorag{} & .538 & .455 & .485 & .540 & .421$^\dagger$ & .362 & .519$^\dagger$ & .452$^\dagger$ & .546$^\dagger$ & .547 & .519$^\dagger$ & .468 & .621$^\dagger$ & .623$^\dagger$ & .609$^\dagger$ & .650$^\dagger$ & .526$^\dagger$ & .521$^\dagger$ \\

\bertgenrag{} & .576$^\dagger$ & .465 & .440 & .713$^\dagger$ & .435 & .359 & .564$^\dagger$ & .390 & .445 & .674$^\dagger$ & .483 & .374 & .631 & .471 & .432 & .643 & .474 & .440 \\

\bertgen{} & \textbf{.713}$^\dagger$ & \textbf{.528}$^\dagger$ & \textbf{.503}$^\dagger$ & \textbf{.788}$^\dagger$ & \textbf{.661}$^\dagger$ & \textbf{.451}$^\dagger$ & \textbf{.733}$^\dagger$ & \textbf{.562}$^\dagger$ & \textbf{.599}$^\dagger$ & \textbf{.779}$^\dagger$ & \textbf{.697}$^\dagger$ & \textbf{.521}$^\dagger$ & \textbf{.873}$^\dagger$ & \textbf{.694}$^\dagger$ & \textbf{.653}$^\dagger$ & \textbf{.868}$^\dagger$ & \textbf{.655}$^\dagger$ & \textbf{.554}$^\dagger$ \\

\midrule
\multicolumn{19}{c}{\color{gray}\textbf{E5 Retrieval}} \\
\midrule
\idf{} & .046 & .015 & .016 & -.045 & -.029 & -.025 & -.042 & -.001 & -.024 & -.041 & .009 & .003 & -.081 & -.063 & -.043 & -.021 & -.013 & .005 \\
\scq{} & .027 & -.002 & .008 & -.043 & -.024 & -.035 & .013 & .023 & .044 & .038 & .037 & .055 & -.053 & -.034 & -.016 & -.004 & .005 & .014 \\
\var{} & .055 & .023 & .036 & -.041 & -.030 & -.012 & -.038 & .007 & -.010 & -.033 & .017 & -.009 & -.058 & -.041 & -.022 & -.011 & -.011 & .001 \\

\midrule
\nqc{} & .167 & .171 & .203 & .208 & .153 & .141 & .104 & .087 & .124 & .159 & .086 & .068 & .141 & .161 & .175 & .159 & .142 & .117 \\
\qc{} & .157 & .163 & .194 & .204 & .150 & .138 & .102 & .086 & .125 & .160 & .089 & .069 & .134 & .154 & .167 & .155 & .140 & .114 \\
\wig{} & .097 & .072 & .084 & .099 & .071 & .067 & -.027 & .006 & .033 & .008 & -.012 & -.018 & -.054 & -.059 & -.015 & .019 & -.007 & -.019 \\
\uWIG{} & .142 & .133 & .157 & .189 & .124 & .097 & .118 & .096 & .144 & .176 & .065 & .063 & .016 & -.025 & .004 & .109 & .067 & .044 \\
\smv{} & .163 & .164 & .196 & .204 & .148 & .136 & .050 & .077 & .117 & .180 & .115 & .105 & .138 & .157 & .173 & .050 & .058 & .053 \\
\uSMV{} & .154 & .156 & .187 & .201 & .146 & .134 & .049 & .076 & .118 & .181 & .118 & .106 & .117 & .116 & .130 & .049 & .057 & .051 \\
\reference{} & .149 & .158 & .178 & .106 & .081 & .071 & .124 & .101 & .143 & .167 & .099 & .101 & .105 & .094 & .115 & .161 & .127 & .126 \\
\bertpost{} & .499$^\dagger$ & .428$^\dagger$ & .481$^\dagger$ & .390$^\dagger$ & .210 & .244$^\dagger$ & .370$^\dagger$ & .328$^\dagger$ & .378$^\dagger$ & .486$^\dagger$ & .233$^\dagger$ & .267$^\dagger$ & .460$^\dagger$ & .456$^\dagger$ & .445$^\dagger$ & .296$^\dagger$ & .282$^\dagger$ & .298$^\dagger$ \\

\siamesereg{} & .606$^\dagger$ & .509$^\dagger$ & .543$^\dagger$ & .539$^\dagger$ & .350$^\dagger$ & .394$^\dagger$ & .484$^\dagger$ & .412$^\dagger$ & .461$^\dagger$ & .570$^\dagger$ & .367$^\dagger$ & .421$^\dagger$ & .565$^\dagger$ & .558$^\dagger$ & .540$^\dagger$ & .485$^\dagger$ & .402$^\dagger$ & .429$^\dagger$ \\

\midrule
\entropy{} & .503 & .355 & .374 & .335 & .008 & .135 & .358 & .198 & .245 & .385 & -.035 & .134 & .564 & .474 & .427 & .512 & .236 & .254 \\

\nli{} & .330 & .294 & .348 & .206 & .157 & .216 & .247 & .226 & .317 & .292 & .156 & .248 & .418 & .493 & .493 & .352 & .313 & .378 \\

\bertgennorag{} & .577 & .562$^\dagger$ & .582$^\dagger$ & .426 & .462$^\dagger$ & .402 & .513 & .475$^\dagger$ & .536$^\dagger$ & .570 & .562$^\dagger$ & .454 & .672$^\dagger$ & .687$^\dagger$ & .671$^\dagger$ & .445 & .561$^\dagger$ & .452 \\

\bertgenrag{} & .654$^\dagger$ & .510 & .509 & .719$^\dagger$ & .434 & .295 & .557$^\dagger$ & .409 & .407 & .676$^\dagger$ & .443 & .326 & .655 & .494 & .465 & .614$^\dagger$ & .385 & .385 \\

\bertgen{} & \textbf{.765}$^\dagger$ & \textbf{.620}$^\dagger$ & \textbf{.611}$^\dagger$ & \textbf{.793}$^\dagger$ & \textbf{.597}$^\dagger$ & \textbf{.502}$^\dagger$ & \textbf{.744}$^\dagger$ & \textbf{.609}$^\dagger$ & \textbf{.590}$^\dagger$ & \textbf{.780}$^\dagger$ & \textbf{.715}$^\dagger$ & \textbf{.523}$^\dagger$ & \textbf{.889}$^\dagger$ & \textbf{.738}$^\dagger$ & \textbf{.704}$^\dagger$ & \textbf{.885}$^\dagger$ & \textbf{.664}$^\dagger$ & \textbf{.642}$^\dagger$ \\

\bottomrule
\end{tabular}

\end{table*}

\subsubsection{{\bf Pre-retrieval Predictors.}}
The pre-retrieval predictors (\idf{}, \scq{}, \var{}
) show low to negligible correlations (approximately $0$-$0.1$) with actual RAG gain across all configurations. 

\subsubsection{{\bf Post-retrieval Predictors.}}
Post-retrieval predictors show a clear improvement over pre-retrieval methods. The unsupervised methods (\nqc{}, \wig{}, \smv{}
) which analyze score distribution, achieve low correlations in the range of approximately $0.1$-$0.2$.
\reference{} outperforms the score-distribution-based predictors in most configurations, 
attaining relatively descent correlations (e.g., $0.25$-$0.30$ on \nqopen{}, for the two LLMs with BM25 retrieval). 
Its superiority over other post-retrieval predictors is more pronounced with BM25 retrieval than with E5 retrieval, as the re-ranker used to construct the reference list is substantially more effective in the former case than in the latter.

The supervised \bertpost{} predictor demonstrates substantially stronger performance. Trained explicitly to predict RAG gain, this model achieves correlations of up to $0.45$-$0.49$ on \nqopen{}. Moreover, \bertpost{} outperforms the REF baseline and all other pre- and post-retrieval predictors by a statistically significant margin in most configurations (34 out of 36). These results highlight the advantage of supervised approaches that capture linguistic and semantic relationships between the question and context that are closely tied to answer quality, beyond what can be achieved with simple score‑based predictors.

\subsubsection{{\bf Post-Generation Predictors.}}


The best performance in Table \ref{tab:combined_results} is attained by some of the post-generation predictors.

\begin{itemize}[leftmargin=0pt]
\item \entropy{} and \nli{}. The gap in uncertainty between the RAG-based answer and the no-RAG answer serves as a strong  signal for model performance, achieving correlations of $0.35$--$0.55$ across most configurations. In contrast, \nli{} exhibits mixed results, performing well in some configurations (e.g., TriviaQA with BM25 retrieval) but poorly in others. This performance gap can be attributed to the fact that \nli{} does not utilize the LLM’s internal knowledge, whereas \entropy{} implicitly captures it by analyzing the model’s output probability distribution.

\item \bertgengeneral{}. These supervised post‑generation predictors, which incorporate only a single generated answer, outperform \bertpost{}, with statistically significance difference in most cases, by leveraging the additional answer information available in their input. 
No clear winner emerges between the two variants, as their relative performance varies across configurations. Moreover, both are statistically significantly inferior to \bertgen{}, which has access to both generated answers.
    
\item{\bertgen{}}. This predictor incorporates the two generated answers alongside the question and context (retrieved passages). It is statistically significantly better than all other predictors, across all configurations, achieving the highest correlations overall, peaking at $0.87-0.89$ on \trivia{} and $0.71-0.79$ on \nqopen{}.

We note that the clear superiority of \bertgen{} over all other prediction methods, as well as over gain prediction results reported in prior work \cite{tian2026predictingretrievalutilityanswer}, establishes it --- to the best of our knowledge --- as the state-of-the-art method for predicting RAG gain.



\end{itemize}

\subsection{{\bf \siamesereg{}} performance}
\label{internal_prober_performance}
\siamesereg{} 
helps to ameliorate the huge performance gap between post-retrieval and post-generation predictors.
Specifically, it posts performance that statistically significantly transcends that of \bertpost{}, and is (often substantially) better than that of three post-generation predictors: \entropy, \nli{} and \bertgengeneral{}.
While it is statistically significantly outperformed by \bertgen{}, it avoids auto-regressive decoding of both candidate answers, therefore reducing generated-token computation by 82–95\%, averaged across datasets and LLMs. 
This predictor provides a practical alternative, in particular for latency-sensitive question answering systems. 

\begin{figure}[ht]
    \centering
    \includegraphics[width=0.7\textwidth]{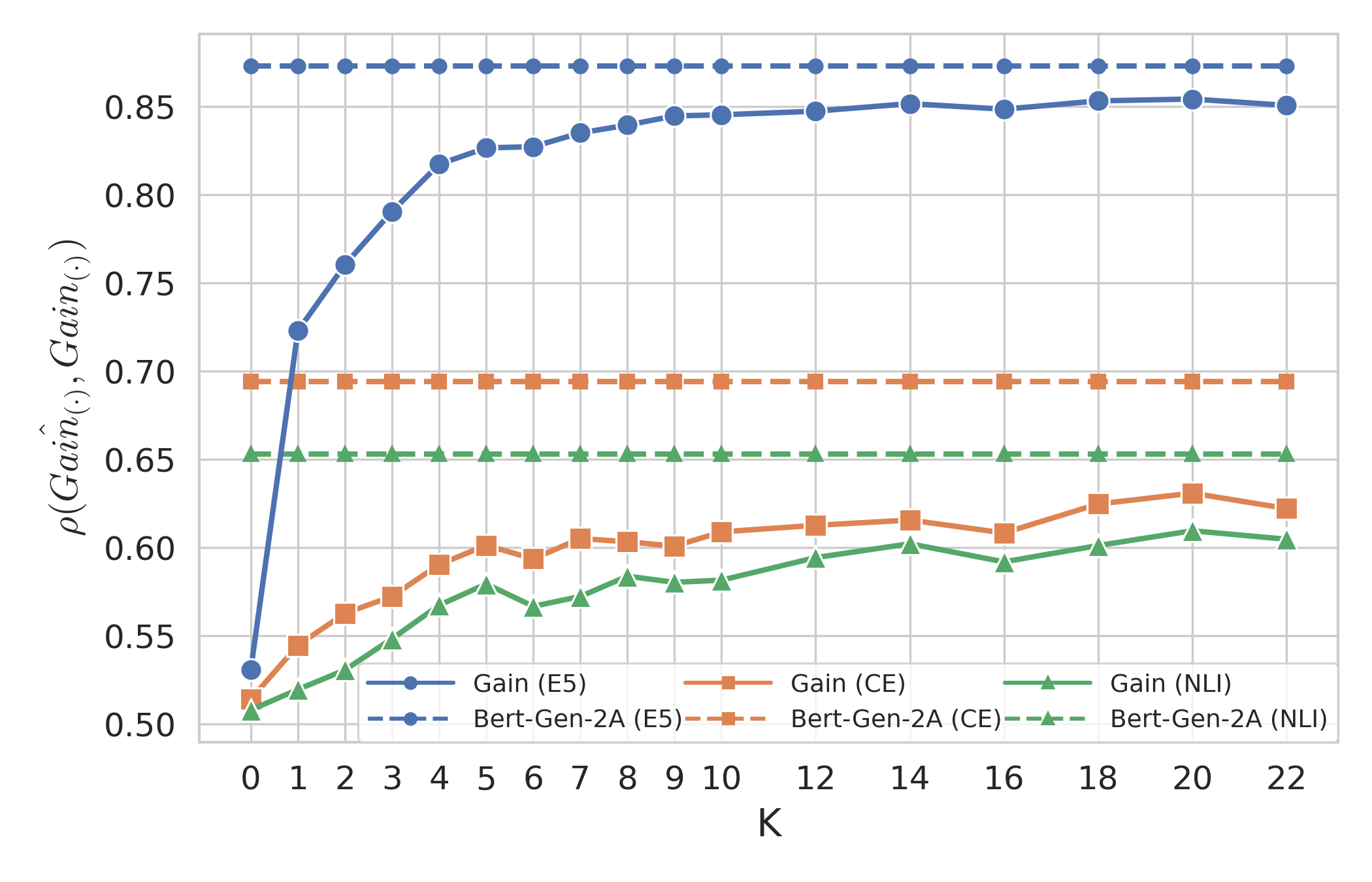}
    \caption{\small
    Pearson correlation between the predicted gain of \siamesereg{} and actual gain, for the three quality metrics, across different values of $K$, the first tokens generated prior to probing. 
    Answers were generated by Falcon with BM25 retrieval across The \trivia{} test set.
    The horizontal lines represent \bertgen{} performance for each of the metrics, measured for the full generated answers,
    which serve as 
    a reference for \siamesereg{} performance.}
    
        
\label{fig:k-tokens}
\end{figure}

We next study the effect of the partial generation employed by \siamesereg{} on prediction quality. Specifically, we allow 
the LLM to generate both RAG and no-RAG answers for a limited number of tokens, and then train \siamesereg{} on these intermediate outputs. Figure~\ref{fig:k-tokens} shows the Pearson correlation between predicted and actual gain as a function of $K$, the number of tokens generated prior to probing. We experiment with Falcon with BM25 retrieval over the \trivia{} test set. We observe a clear upward trend: even a small amount of token generation substantially improves prediction quality, with performance approaching that of \bertgen{} as $K$ increases.
Notably, the largest increase occurs within the first generated tokens, indicating that early generation already provides strong signals for gain estimation. The performance continues to improve more gradually and saturates around $K=8$ tokens.

\subsection{Cross-Architecture and Cross-Metric Consistency}
The 
correlation trends observed in
Table \ref{tab:combined_results}
demonstrate {high consistency} across experimental settings. The relative ranking of predictors remains largely unchanged when switching between retrieval methods (BM25 and E5). Similarly, consistent trends are observed for both LLMs, Falcon and Llama, indicating robustness to LLM architecture. Moreover, 
the ordering of predictors based on correlations are the same for the different answer-quality 
 metrics, albeit 
 in most cases the observed correlation numbers of
 \accefive{} are  higher than those for \accce{} and \accnli{}. 
 This finding may be attributed to the observation that the \accefive{} cosine values are empirically distributed relatively smoothly over the range $[0.7,1]$, a well-known property of the E5 embedder. By contrast, \accce{} and \accnli{} values span the full range $[0,1]$ with high variability, when values pile up near the bounds, thereby degrading correlation between predicted and actual gains. See Figure \ref{fig:gain_distributions} for the differences is actual gain distribution between the three metrics.



\section{Selective RAG}
\label{selective_rag}


Given the strong gain‑prediction performance 
reported above,
we evaluate the effectiveness of some of our best predictors (\bertpost{}, \siamesereg{}, \bertgen{}) in a selective RAG setting. The objective is to apply RAG on a per‑question basis only when an improvement in answer quality is expected. Our selective RAG strategies follow a threshold‑based policy over the predicted gain $\hat{G}$; a positive predicted gain indicates that the RAG‑based answer is expected to outperform the no‑RAG alternative (i.e., the quality ratio exceeds one), whereas a non‑positive gain suggests that RAG is likely to degrade performance. Accordingly, when $\hat{G} >0$, we generate the answer using RAG; otherwise, we invoke the LLM without retrieval to produce a no‑RAG response.

\paragraph{{\bf Evaluation}}
We evaluate our selective RAG approach by measuring the average quality of the selected answers over the 
test sets
of each of the three Q\&A datasets. Higher average quality indicates more effective selection decisions. Answer quality is assessed using one of the three quality metrics defined in Section~\ref{subsec: quality-metrics}. For each selector, evaluation is conducted using the same quality metric that was employed to train the corresponding gain predictor. For example, selectors based on $\hat{G}_{CE}$ are evaluated by averaging $\mathcal{Q}_{CE}$ over selected answers across all questions in the test set.
In addition, we report the accuracy of each selector; i.e., the portion of questions for which the decision of applying RAG (or not) was correct. 

We compare our predicted‑gain–based selectors against three baseline strategies:
\begin{itemize}[leftmargin=0pt]
\item \emph{No RAG}: Always select the no‑RAG answer.
\item \emph{Always RAG}: Always select the RAG‑based answer.
\item \emph{Oracle}: For each question, select the higher‑quality answer between the no‑RAG and RAG‑based responses. This optimal selection provides an upper bound on the performance achievable by gain-prediction-based selection methods.
\end{itemize}

We determine statistical significant difference between selectors's performance using the 
paired two-tailed 
t-test with $p=0.05$,
and with Bonferroni correction for multiple hypothesis testing.

\paragraph{{\bf Results}}
Table~\ref{tab:selective_rag} reports the 
performance of the different selection approaches.
The first observation is that selectors based on  $\hat{G}_{E5}$ consistently exhibit higher 
answer-quality scores than those of the other answer-quality metrics, although the accuracy is not necessarily higher. This finding can be explained, as noted above, by the range of scores ($[0.7-1.0]$) produced by this metric\footnote{\url{https://huggingface.co/intfloat/e5-large-v2}}.
Consequently, answer-quality performance should be interpreted in terms of improvements relative to the baselines rather than absolute score magnitudes.
Accordingly, for each configuration and for each selection method $M$ with quality performance $Q(M)$, we measure its relative performance gap (in percentage) with respect to the \emph{Always RAG} and \emph{Oracle} baselines by $\frac{Q(M) - Q(AlwaysRAG)}{Q(Oracle)-Q(AlwaysRAG)}\cdot 100$.

\emph{Always RAG} 
statistically
significantly outperforms \emph{No RAG}, 
which attests to 
effectiveness of RAG for question answering. Nevertheless, a substantial performance gap remains between \emph{Always RAG} and \emph{Oracle}, a gap which prediction‑based selectors aim to reduce.

Across nearly all configurations, all prediction-based selectors consistently outperform \textit{Always RAG} in terms of answer-quality and accuracy. Among these selectors, \bertgen{} delivers the strongest and most consistent improvements, achieving statistically significant gains over the baselines in almost all settings. It also frequently recovers a substantial portion of the performance gap relative to \emph{Oracle}. 
These results confirm that selective RAG based on gain prediction provides an effective mechanism for controlling RAG usage while maintaining high answer quality.

The \bertpost{}-based selector rarely attains the highest performance among selectors, however, it consistently improves upon \emph{Always RAG} and exhibits more stable behavior across retrievers.
The \siamesereg{} model performs competitively, often matching or exceeding \bertgen{} under the $\mathcal{Q}_{NLI}$ metric, in line with its strong gain‑prediction capability. 
However, its performance is less consistent across retrievers, indicating increased sensitivity to retrieval noise. 

A notable trend is that selective RAG approaches generally yield larger relative improvements when paired with BM25 retrieval than with E5‑based retrieval. This observation aligns with the noisier nature of lexical retrieval, where selectively filtering out distracting context for augmentation plays a more critical role.

\begin{table*}[t]
\centering
\caption{\small
Selective RAG performance in terms of answer quality and accuracy.
{\em  Gap} is the portion of the performance gap between {\em Always RAG} and {\em Oracle}.
Statistically significant differences with Always RAG and No RAG are marked by '$a$' and '$n$', respectively; $'g'$ marks a statistically significant difference with {\em all} the selectors above the current selector. Boldface marks the best performance per column for a retrieval method..
}
\resizebox{\textwidth}{!}{
 \begin{tabular}{l||ccc|ccc||ccc|ccc||ccc|ccc}
\toprule
& \multicolumn{6}{c||}{\textbf{\nqopen{}}} & \multicolumn{6}{c||}{\textbf{\hotpot{}}} & \multicolumn{6}{c}{\textbf{\trivia{}}} \\
\midrule
  & \multicolumn{3}{c|}{\textbf{Falcon}} & \multicolumn{3}{c||}{\textbf{Llama}} & \multicolumn{3}{c|}{\textbf{Falcon}} & \multicolumn{3}{c||}{\textbf{Llama}} & \multicolumn{3}{c|}{\textbf{Falcon}} & \multicolumn{3}{c}{\textbf{Llama}} \\
Selector & \accefive{} & \accce{} & \accnli{} & \accefive{} & \accce{} & \accnli{} & \accefive{} & \accce{} & \accnli{} & \accefive{} & \accce{} & \accnli{} & \accefive{} & \accce{} & \accnli{} & \accefive{} & \accce{} & \accnli{} \\ \\
\midrule
\multicolumn{19}{c}{\color{gray}\textbf{BM25 Retrieval}} \\
\midrule
No RAG: & & & & & & & & & & & & & & & & & & \\
\textit{Quality} & .786 & .221 & .309 & .795 & .306 & .439 & .780 & .188 & .230 & .776 & .251 & .277 & .862 & .474 & .607 & .880 & .566 & .680 \\
\textit{Accuracy} & 52.0\% & 42.2\% & 43.9\% & 55.3\% & 48.3\% & 57.5\% & 41.6\% & 37.3\% & 37.9\% & 40.7\% & 42.0\% & 50.8\% & 51.1\% & 50.7\% & 52.8\% & 50.0\% & 50.3\% & 64.7\% \\[0.25ex]
Always RAG: & & & & & & & & & & & & & & & & & & \\
\textit{Quality} & .796 & .284 & .387 & .803 & .345 & .329 & .812 & .331 & .386 & .816 & .360 & .305 & .891 & .591 & .706 & .911 & .670 & .636 \\
\textit{Accuracy} & 51.4\% & 60.4\% & 58.3\% & 46.6\% & 53.5\% & 44.2\% & 63.8\% & 67.3\% & 67.9\% & 61.6\% & 60.0\% & 50.3\% & 65.7\% & 70.2\% & 60.2\% & 67.6\% & 70.5\% & 45.2\% \\[0.25ex]
Oracle: & & & & & & & & & & & & & & & & & & \\
\textit{Quality} & .813 & .339 & .476 & .830 & .448 & .538 & .828 & .376 & .443 & .833 & .443 & .428 & .910 & .658 & .776 & .936 & .759 & .779 \\
\midrule
\bertpost{}: & & & & & & & & & & & & & & & & & & \\
\textit{Quality} & $.801^{\mathrm{a}}_{\mathrm{n}}$ & $.284_{\mathrm{n}}$ & $.408^{\mathrm{a}}_{\mathrm{n}}$ & $.814^{\mathrm{a}}_{\mathrm{n}}$ & $.367^{\mathrm{a}}_{\mathrm{n}}$ & $.442^{\mathrm{a}}$ & $.812_{\mathrm{n}}$ & $.332_{\mathrm{n}}$ & $.391_{\mathrm{n}}$ & $.819^{\mathrm{a}}_{\mathrm{n}}$ & $.369^{\mathrm{a}}_{\mathrm{n}}$ & $.342^{\mathrm{a}}_{\mathrm{n}}$ & $.891_{\mathrm{n}}$ & $.594_{\mathrm{n}}$ & $.713^{\mathrm{a}}_{\mathrm{n}}$ & $.911_{\mathrm{n}}$ & $.684^{\mathrm{a}}_{\mathrm{n}}$ & $.699^{\mathrm{a}}_{\mathrm{n}}$ \\
\textit{Gap} & \textit{$\uparrow$29.4\%} & \textit{0.0\%} & \textit{$\uparrow$23.6\%} & \textit{$\uparrow$40.7\%} & \textit{$\uparrow$21.4\%} & \textit{$\uparrow$54.1\%} & \textit{0.0\%} & \textit{$\uparrow$2.2\%} & \textit{$\uparrow$8.8\%} & \textit{$\uparrow$17.6\%} & \textit{$\uparrow$10.8\%} & \textit{$\uparrow$30.1\%} & \textit{0.0\%} & \textit{$\uparrow$4.5\%} & \textit{$\uparrow$10.0\%} & \textit{0.0\%} & \textit{$\uparrow$15.7\%} & \textit{$\uparrow$44.1\%} \\
\textit{Accuracy} & 63.6\% &  60.3\% &  58.7\% & 64.1\% &  60.5\% &  58.4\% &  64.1\% & 67.4\% & 67.4\% & 67.0\% & 62.9\% & 58.7\% &  65.8\% & 71.1\% & 62.9\% & 67.6\% & 73.1\% & 58.8\% \\[1ex]

\siamesereg{}: & & & & & & & & & & & & & & & & & & \\
\textit{Quality} & $.803^{\mathrm{a}\mathrm{g}}_{\mathrm{n}}$ & $.294^{\mathrm{a}\mathrm{g}}_{\mathrm{n}}$ & $\mathbf{.414}^{\mathrm{a}}_{\mathrm{n}}$ & $.819^{\mathrm{a}\mathrm{g}}_{\mathrm{n}}$ & $.380^{\mathrm{a}\mathrm{g}}_{\mathrm{n}}$ & $\mathbf{.455}^{\mathrm{a}\mathrm{g}}_{\mathrm{n}}$ & $.815^{\mathrm{a}\mathrm{g}}_{\mathrm{n}}$ & $.333_{\mathrm{n}}$ & $.388_{\mathrm{n}}$ & $.821^{\mathrm{a}\mathrm{g}}_{\mathrm{n}}$ & $.377^{\mathrm{a}\mathrm{g}}_{\mathrm{n}}$ & $.354^{\mathrm{a}\mathrm{g}}_{\mathrm{n}}$ & $.897^{\mathrm{a}\mathrm{g}}_{\mathrm{n}}$ & $.604^{\mathrm{a}\mathrm{g}}_{\mathrm{n}}$ & $.724^{\mathrm{a}\mathrm{g}}_{\mathrm{n}}$ & $.922^{\mathrm{a}\mathrm{g}}_{\mathrm{n}}$ & $.703^{\mathrm{a}\mathrm{g}}_{\mathrm{n}}$ & $.712^{\mathrm{a}\mathrm{g}}_{\mathrm{n}}$ \\

\textit{Gap} & \textit{$\uparrow$41.2\%} & \textit{$\uparrow$18.2\%} & \textbf{\textit{$\uparrow$30.3\%}} & \textit{$\uparrow$59.3\%} & \textit{$\uparrow$34.0\%} & \textbf{\textit{$\uparrow$60.3\%}} & \textit{$\uparrow$18.7\%} & \textit{$\uparrow$4.4\%} & \textit{$\uparrow$3.5\%} & \textit{$\uparrow$29.4\%} & \textit{$\uparrow$20.5\%} & \textit{$\uparrow$39.8\%} & \textit{$\uparrow$31.6\%} & \textit{$\uparrow$19.4\%} & \textit{$\uparrow$25.7\%} & \textit{$\uparrow$44.0\%} & \textit{$\uparrow$37.1\%} & \textit{$\uparrow$53.1\%} \\
\textit{Accuracy} & 64.9\% & 62.0\% & \textbf{60.2\%} & 69.8\% & 63.1\% & 59.9\% & 66.7\% & 67.1\% & 66.4\% & 68.9\% & 63.7\% & 60.6\% & 73.3\% & 73.4\% & 63.3\% & 76.6\% &  75.6\% & 59.2\% \\[1ex]

\bertgen{}: & & & & & & & & & & & & & & & & & & \\
\textit{Quality} & $\mathbf{.806}^{\mathrm{a}\mathrm{g}}_{\mathrm{n}}$ & $\mathbf{.296}^{\mathrm{a}}_{\mathrm{n}}$ & $.410^{\mathrm{a}}_{\mathrm{n}}$ & $\mathbf{.824}^{\mathrm{a}\mathrm{g}}_{\mathrm{n}}$ & $\mathbf{.408}^{\mathrm{a}\mathrm{g}}_{\mathrm{n}}$ & $.451^{\mathrm{a}}_{\mathrm{n}}$ & $\mathbf{.820}^{\mathrm{a}\mathrm{g}}_{\mathrm{n}}$ & $\mathbf{.341}^{\mathrm{a}\mathrm{g}}_{\mathrm{n}}$ & $\mathbf{.399}^{\mathrm{a}\mathrm{g}}_{\mathrm{n}}$ & $\mathbf{.826}^{\mathrm{a}\mathrm{g}}_{\mathrm{n}}$ & $\mathbf{.406}^{\mathrm{a}\mathrm{g}}_{\mathrm{n}}$ & $\mathbf{.372}^{\mathrm{a}\mathrm{g}}_{\mathrm{n}}$ & $\mathbf{.906}^{\mathrm{a}\mathrm{g}}_{\mathrm{n}}$ & $\mathbf{.629}^{\mathrm{a}\mathrm{g}}_{\mathrm{n}}$ & $\mathbf{.730}^{\mathrm{a}}_{\mathrm{n}}$ & $\mathbf{.932}^{\mathrm{a}\mathrm{g}}_{\mathrm{n}}$ & $\mathbf{.729}^{\mathrm{a}\mathrm{g}}_{\mathrm{n}}$ & $\mathbf{.730}^{\mathrm{a}\mathrm{g}}_{\mathrm{n}}$ \\
\textit{Gap} & \textbf{\textit{$\uparrow$58.8\%}} & \textbf{\textit{$\uparrow$21.8\%}} & \textit{$\uparrow$25.8\%} & \textbf{\textit{$\uparrow$77.8\%}} & \textbf{\textit{$\uparrow$61.2\%}} & \textit{$\uparrow$58.4\%} & \textbf{\textit{$\uparrow$50.0\%}} & \textbf{\textit{$\uparrow$22.2\%}} & \textbf{\textit{$\uparrow$22.8\%}} & \textbf{\textit{$\uparrow$58.8\%}} & \textbf{\textit{$\uparrow$55.4\%}} & \textbf{\textit{$\uparrow$54.5\%}} & \textbf{\textit{$\uparrow$78.9\%}} & \textbf{\textit{$\uparrow$56.7\%}} & \textbf{\textit{$\uparrow$34.3\%}} & \textbf{\textit{$\uparrow$84.0\%}} & \textbf{\textit{$\uparrow$66.3\%}} & \textbf{\textit{$\uparrow$65.7\%}} \\
\textit{Accuracy} & \textbf{71.2\%} & \textbf{63.2\%} & 59.2\% & \textbf{77.6\%} & \textbf{72.6\%} & \textbf{62.3\%} & \textbf{72.7\%} & \textbf{69.5\%} & \textbf{69.3\%} & \textbf{77.0\%} & \textbf{74.3\%} & \textbf{67.0\%} & \textbf{83.2\%} & \textbf{79.8\%} & \textbf{68.8\%} & \textbf{87.1\%} & \textbf{82.3\%} & \textbf{71.1\%} \\

\midrule
\multicolumn{19}{c}{\color{gray}\textbf{E5 Retrieval}} \\
\midrule
No RAG: & & & & & & & & & & & & & & & & & & \\
\textit{Quality} & .786 & .221 & .309 & .795 & .307 & .442 & .780 & .188 & .230 & .775 & .249 & .277 & .862 & .474 & .607 & .880 & .564 & .678 \\
\textit{Accuracy} & 39.9\% & 34.3\% & 36.3\% & 38.3\% & 38.8\% & 50.1\% & 37.7\% & 34.5\% & 37.0\% & 34.9\% & 37.7\% & 47.2\% & 45.9\% & 46.4\% & 50.6\% & 44.8\% & 46.9\% & 61.4\% \\[0.25ex]
Always RAG: & & & & & & & & & & & & & & & & & & \\
\textit{Quality} & .816 & .355 & .536 & .833 & .424 & .497 & .819 & .355 & .423 & .826 & .391 & .362 & .902 & .632 & .752 & .927 & .725 & .706 \\
\textit{Accuracy} & 64.0\% & 68.7\% & 66.1\% & 63.4\% & 63.1\% & 51.9\% & 67.3\% & 70.4\% & 69.4\% & 67.5\% & 64.8\% & 54.3\% & 70.1\% & 72.7\% & 61.5\% & 74.2\% & 75.1\% & 50.4\% \\[0.25ex]
Oracle: & & & & & & & & & & & & & & & & & & \\
\textit{Quality} & .828 & .392 & .586 & .847 & .491 & .611 & .833 & .393 & .468 & .838 & .459 & .455 & .919 & .691 & .808 & .942 & .784 & .797 \\

\midrule
\bertpost{}: & & & & & & & & & & & & & & & & & & \\
\textit{Quality} & $.817^{\mathrm{a}}_{\mathrm{n}}$ & $.355_{\mathrm{n}}$ & $.534_{\mathrm{n}}$ & $.833_{\mathrm{n}}$ & $.423_{\mathrm{n}}$ & $.502_{\mathrm{n}}$ & $.818_{\mathrm{n}}$ & $.354_{\mathrm{n}}$ & $.425_{\mathrm{n}}$ & $.826_{\mathrm{n}}$ & $.388_{\mathrm{n}}$ & $.369^{\mathrm{a}}_{\mathrm{n}}$ & $.904^{\mathrm{a}}_{\mathrm{n}}$ & $.635_{\mathrm{n}}$ & $.754_{\mathrm{n}}$ & $.927_{\mathrm{n}}$ & $.719_{\mathrm{n}}$ & $.720^{\mathrm{a}}_{\mathrm{n}}$ \\

\textit{Gap} & \textit{$\uparrow$8.3\%} & \textit{0.0\%} & \textit{$\downarrow$4.0\%} & \textit{0.0\%} & \textit{$\downarrow$1.5\%} & \textit{$\uparrow$4.4\%} & \textit{$\downarrow$7.1\%} & \textit{$\downarrow$2.6\%} & \textit{$\uparrow$4.4\%} & \textit{0.0\%} & \textit{$\downarrow$4.4\%} & \textit{$\uparrow$7.5\%} & \textit{$\uparrow$11.8\%} & \textit{$\uparrow$5.1\%} & \textit{$\uparrow$3.6\%} & \textit{0.0\%} & \textit{$\downarrow$10.2\%} & \textit{$\uparrow$15.4\%} \\
\textit{Accuracy} & 66.9\% & 68.7\% & 65.8\% & 64.3\% & 63.3\% & 54.7\% & 67.0\% & 70.3\% & 69.1\% & 69.8\% & 64.3\% & 57.0\% & 73.4\% & 73.6\% & 62.6\% & 73.9\% & 73.6\% & 56.9\% \\[1ex]

\siamesereg{}: & & & & & & & & & & & & & & & & & & \\
\textit{Quality} & $.818^{\mathrm{a}\mathrm{g}}_{\mathrm{n}}$ & $.355_{\mathrm{n}}$ & $\mathbf{.536}_{\mathrm{n}}$ & $.835^{\mathrm{a}\mathrm{g}}_{\mathrm{n}}$ & $.429^{\mathrm{a}\mathrm{g}}_{\mathrm{n}}$ & $\mathbf{.531}^{\mathrm{a}\mathrm{g}}_{\mathrm{n}}$ & $.819_{\mathrm{n}}$ & $.354_{\mathrm{n}}$ & $.421_{\mathrm{n}}$ & $.828^{\mathrm{a}\mathrm{g}}_{\mathrm{n}}$ & $.396^{\mathrm{g}}_{\mathrm{n}}$ & $.384^{\mathrm{a}\mathrm{g}}_{\mathrm{n}}$ & $.905^{\mathrm{a}}_{\mathrm{n}}$ & $.638^{\mathrm{a}}_{\mathrm{n}}$ & $.757^{\mathrm{a}}_{\mathrm{n}}$ & $.930^{\mathrm{a}\mathrm{g}}_{\mathrm{n}}$ & $.733^{\mathrm{a}\mathrm{g}}_{\mathrm{n}}$ & $.736^{\mathrm{a}\mathrm{g}}_{\mathrm{n}}$ \\

\textit{Gap} & \textit{$\uparrow$16.7\%} & \textit{0.0\%} & \textbf{\textit{0.0\%}} & \textit{$\uparrow$14.3\%} & \textit{$\uparrow$7.5\%} & \textbf{\textit{$\uparrow$29.8\%}} & \textit{0.0\%} & \textit{$\downarrow$2.6\%} & \textit{$\downarrow$4.4\%} & \textit{$\uparrow$16.7\%} & \textit{$\uparrow$7.4\%} & \textit{$\uparrow$23.7\%} & \textit{$\uparrow$17.6\%} & \textit{$\uparrow$10.2\%} & \textit{$\uparrow$8.9\%} & \textit{$\uparrow$20.0\%} & \textit{$\uparrow$13.6\%} & \textit{$\uparrow$33.0\%} \\
\textit{Accuracy} & 68.6\% & 68.5\% & 66.1\% & 68.3\% & 63.9\% & 57.1\% & 68.4\% & 69.5\% & 67.8\% & 71.5\% & 64.5\% & 59.9\% & 73.9\% & 74.3\% & 63.4\% & 76.5\% & 77.1\% & 58.2\% \\[1ex]

\bertgen{}: & & & & & & & & & & & & & & & & & & \\
\textit{Quality} & $\mathbf{.822}^{\mathrm{a}\mathrm{g}}_{\mathrm{n}}$ & $\mathbf{.362}^{\mathrm{a}\mathrm{g}}_{\mathrm{n}}$ & $.533_{\mathrm{n}}$ & $\mathbf{.840}^{\mathrm{a}\mathrm{g}}_{\mathrm{n}}$ & $\mathbf{.451}^{\mathrm{a}\mathrm{g}}_{\mathrm{n}}$ & $.526^{\mathrm{a}}_{\mathrm{n}}$ & $\mathbf{.825}^{\mathrm{a}\mathrm{g}}_{\mathrm{n}}$ & $\mathbf{.366}^{\mathrm{a}\mathrm{g}}_{\mathrm{n}}$ & $\mathbf{.431}^{\mathrm{a}\mathrm{g}}_{\mathrm{n}}$ & $\mathbf{.832}^{\mathrm{a}\mathrm{g}}_{\mathrm{n}}$ & $\mathbf{.428}^{\mathrm{a}\mathrm{g}}_{\mathrm{n}}$ & $\mathbf{.410}^{\mathrm{a}\mathrm{g}}_{\mathrm{n}}$ & $\mathbf{.916}^{\mathrm{a}\mathrm{g}}_{\mathrm{n}}$ & $\mathbf{.667}^{\mathrm{a}\mathrm{g}}_{\mathrm{n}}$ & $\mathbf{.773}^{\mathrm{a}\mathrm{g}}_{\mathrm{n}}$ & $\mathbf{.939}^{\mathrm{a}\mathrm{g}}_{\mathrm{n}}$ & $\mathbf{.759}^{\mathrm{a}\mathrm{g}}_{\mathrm{n}}$ & $\mathbf{.754}^{\mathrm{a}\mathrm{g}}_{\mathrm{n}}$ \\

\textit{Gap} & \textbf{\textit{$\uparrow$50.0\%}} & \textbf{\textit{$\uparrow$18.9\%}} & \textit{$\downarrow$6.0\%} & \textbf{\textit{$\uparrow$50.0\%}} & \textbf{\textit{$\uparrow$40.3\%}} & \textit{$\uparrow$25.4\%} & \textbf{\textit{$\uparrow$42.9\%}} & \textbf{\textit{$\uparrow$28.9\%}} & \textbf{\textit{$\uparrow$17.8\%}} & \textbf{\textit{$\uparrow$50.0\%}} & \textbf{\textit{$\uparrow$54.4\%}} & \textbf{\textit{$\uparrow$51.6\%}} & \textbf{\textit{$\uparrow$82.4\%}} & \textbf{\textit{$\uparrow$59.3\%}} & \textbf{\textit{$\uparrow$37.5\%}} & \textbf{\textit{$\uparrow$80.0\%}} & \textbf{\textit{$\uparrow$57.6\%}} & \textbf{\textit{$\uparrow$52.7\%}} \\
\textit{Accuracy} & \textbf{74.8\%} & \textbf{70.9\%} & \textbf{67.3}\% & \textbf{74.7\%} & \textbf{70.5\%} & \textbf{60.5\%} & \textbf{74.4\%} & \textbf{73.6\%} & \textbf{70.0\%} & \textbf{77.9\%} & \textbf{75.4\%} & \textbf{68.3\%} & \textbf{84.4\%} & \textbf{81.5\%} & \textbf{74.5\%} & \textbf{87.6\%} & \textbf{83.0\%} & \textbf{71.7\%} \\[1ex]

\bottomrule
\end{tabular}   
}
\label{tab:selective_rag}

\end{table*}

In our setup, 
each selector is evaluated using the same quality metric on which it was trained. This raises the question of whether the relative ordering
of 
selectors is preserved when all selection methods are evaluated using the same quality metric. Table~\ref{tab:selective_rag_unified} reports the performance of all selectors for the Falcon/BM25 configurations, evaluated using \accllm{} 
(LLM as a judge) 
as a unified measure for answer quality. Interestingly, in this setting, \siamesereg{} outperforms \bertgen{}. This behavior can be attributed to the fact that \siamesereg{} is trained to jointly predict the multiple quality metrics, which likely makes it more robust with respect to other selectors when evaluated under a metric that it was not explicitly optimized for.

\begin{table}[t]
\centering
\scriptsize
\setlength{\tabcolsep}{2pt}
\caption{\small
Selective RAG performance measured by averaging \accllm{} of the selected answers. Falcon is used for answer generation and BM25 is used for retrieval. '$a$' and '$n$' mark statistically significant differences with  Always RAG and No RAG, respectively; '$g$' marks a statistically significant difference with {\em all} selectors above the current one. Boldface: the best result in a column.
}
\begin{tabular}{l||ccc||ccc||ccc}
\toprule
  & \multicolumn{3}{c||}{\textbf{NQ-Open}} & \multicolumn{3}{c||}{\textbf{HotpotQA}} & \multicolumn{3}{c}{\textbf{TriviaQA}} \\
Selector & $Q_{E5}$ & $Q_{CE}$ & $Q_{NLI}$ & $Q_{E5}$ & $Q_{CE}$ & $Q_{NLI}$ & $Q_{E5}$ & $Q_{CE}$ & $Q_{NLI}$ \\
\midrule
\multicolumn{10}{c}{\color{gray}\textbf{Falcon - BM25 Retrieval}} \\
\midrule

No RAG & .380 & .380 & .380 & .317 & .317 & .317 & .623 & .623 & .623 \\
\ \ \ \textit{Accuracy} & 42.3\% & 42.3\% & 42.3\% & 35.0\% & 35.0\% & 35.0\% & 41.7\% & 41.7\% & 41.7\% \\[0.25ex]

Always RAG & .477 & .477 & .477 & .525 & .525 & .525 & .764 & .764 & .764 \\
\ \ \ \textit{Accuracy} & 58.6\% & 58.6\% & 58.6\% & 67.1\% & 67.1\% & 67.1\% & 59.6\% & 59.6\% & 59.6\% \\[0.25ex]

Oracle & .558 & .558 & .558 & .587 & .587 & .587 & .821 & .821 & .821 \\
\midrule

\bertpost{} & $.491^{\mathrm{a}}_{\mathrm{n}}$ & $.477_{\mathrm{n}}$ & $.495^{\mathrm{a}}_{\mathrm{n}}$ & $.519_{\mathrm{n}}$ & $.527_{\mathrm{n}}$ & $.529_{\mathrm{n}}$ & $.765_{\mathrm{n}}$ & $.771^{\mathrm{a}}_{\mathrm{n}}$ & $.770_{\mathrm{n}}$ \\
\ \ \ \textit{Gap} & $\uparrow$16.9\% & $\uparrow$0.2\% & $\uparrow$21.9\% & $\downarrow$8.5\% & $\uparrow$4.0\% & $\uparrow$7.0\% & $\uparrow$1.9\% & $\uparrow$12.5\% & $\uparrow$11.3\% \\
\ \ \ \textit{Accuracy} & 60.2\% & 58.3\% & 61.3\% & 66.6\% & \textbf{68.2\%} & \textbf{68.1\%} & 59.8\% & 60.2\% & 59.9\% \\

\bertgen{} & $.492^{\mathrm{a}}_{\mathrm{n}}$ & $.489^{\mathrm{a}\mathrm{g}}_{\mathrm{n}}$ & $.496^{\mathrm{a}}_{\mathrm{n}}$ & $.513_{\mathrm{n}}$ & $.522_{\mathrm{n}}$ & $\mathbf{.533}^{\mathrm{a}}_{\mathrm{n}}$ & $.773^{\mathrm{a}\mathrm{g}}_{\mathrm{n}}$ & $.779^{\mathrm{a}\mathrm{g}}_{\mathrm{n}}$ & $.778^{\mathrm{a}\mathrm{g}}_{\mathrm{n}}$ \\
\ \ \ \textit{Gap} & $\uparrow$17.9\% & $\uparrow$14.4\% & $\uparrow$23.8\% & $\downarrow$18.8\% & $\downarrow$4.9\% & \textbf{$\uparrow$13.2\%} & $\uparrow$16.9\% & $\uparrow$27.1\% & $\uparrow$25.3\%\\

\ \ \ \textit{Accuracy} & 57.9\% & 61.6\% & 61.6\% & 61.1\% & 66.1\% & 67.8\% & 61.0\% & 60.9\% & 59.9\% \\

\siamesereg{} & $\mathbf{.501}^{\mathrm{a}\mathrm{g}}_{\mathrm{n}}$ & $\mathbf{.496}^{\mathrm{a}\mathrm{g}}_{\mathrm{n}}$ & $\mathbf{.501}^{\mathrm{a}}_{\mathrm{n}}$ & $\mathbf{.530}^{\mathrm{g}}_{\mathrm{n}}$ & $\mathbf{.528}_{\mathrm{n}}$ & $.528_{\mathrm{n}}$ & $\mathbf{.782}^{\mathrm{a}\mathrm{g}}_{\mathrm{n}}$ & $\mathbf{.782}^{\mathrm{a}}_{\mathrm{n}}$ & $\mathbf{.781}^{\mathrm{a}}_{\mathrm{n}}$ \\
\ \ \ \textit{Gap} & \textbf{$\uparrow$29.6\%} & \textbf{$\uparrow$23.8\%} & \textbf{$\uparrow$30.0\%} & \textbf{$\uparrow$8.2\%} & \textbf{$\uparrow$4.9\%} & $\uparrow$5.9\% & \textbf{$\uparrow$31.8\%} &  \textbf{$\uparrow$31.9\%} & \textbf{$\uparrow$31.0\%} \\

\ \ \ \textit{Accuracy} & \textbf{62.3\%} & \textbf{62.1\%} & \textbf{62.7\%} & \textbf{68.1\%} & 67.7\% & 67.3\% & \textbf{61.4\%} & \textbf{61.8\%} & \textbf{60.9\%} \\

\bottomrule
\end{tabular}
\label{tab:selective_rag_unified}
\end{table}


\section{Conclusions and Future Work}

In this work, we addressed the problem of predicting the benefit of applying RAG in LLM‑based question answering. 
We evaluated the effectiveness of several pre‑ and post‑retrieval predictors originally proposed for ad hoc retrieval, as well as a set of post‑generation predictors. 
Our results indicate that the most effective prediction approach is a supervised predictor that explicitly models the semantic relationships among the question, the retrieved passages, and the generated answers. To this end, we introduced \siamesereg{}, a novel predictor that probes the LLM’s internal representations for gain prediction,
offering a practical and computationally efficient alternative to the costly post‑generation methods.

We further evaluated our predictors in a selective RAG setting, in which RAG is applied on a per‑question basis only when an improvement in answer quality is expected. Our experiments demonstrate that selective RAG guided by our leading gain predictors consistently outperforms the strategy of uniformly applying RAG to all questions. 
For future work, we plan to explore additional predictive signals, integrating different predictors, and to evaluate our methods in real‑world question answering scenarios.


Our study has three main limitations. First, \siamesereg{} requires white-box access to the LLM's internal representations and therefore cannot be applied to proprietary black-box models; post-retrieval and generation-based predictors remain applicable in such settings. Second, our evaluation is limited to mid-sized (8--10B parameter) LLMs due to hardware constraints. Finally, \bertgen{} is computationally expensive and is included primarily as a strong reference point for prediction performance rather than as a practical deployment solution.
\balance

\medskip
\noindent {\bf Acknowledgments}
We thank the reviewers for their comments. This work was supported in part by the Israel Science Foundation (ISF grant no. 403/22) and the Technology Innovation Institute (TII). All opinions, conclusions and findings in this paper are those of the authors and do not necessarily represent the views of the ISF or TII.

\bibliographystyle{unsrt}
\bibliography{my-citations}

\appendix
%


\newtcolorbox{promptbox}[1]{breakable=false,left=2pt,right=2pt,fontupper={\small\ttfamily}, fontlower={\small\ttfamily},title={#1},fonttitle={\small},skin=enhanced jigsaw,colframe=black,colback=gray!5!white,size=fbox}



\section{LLM-as-a-Judge prompt}
\label{app:prompts}
\begin{promptbox}{\scriptsize LLM-as-a-Judge Prompt (DeepEval)}
\scriptsize
\textbf{Task:} Evaluate the quality of the \emph{actual answer} with respect to the \emph{expected answer}.

\textbf{Evaluation Criteria:}
\begin{itemize}[leftmargin=0.3cm]
    \item The 'expected answer' contains valid answers separated by '|', treat each alternative as equally valid
    \item Score high if the 'actual answer' matches at least one expected answer after normalization
    \item Penalize if the correct answer is buried in irrelevant content, or if correct and incorrect claims appear together
\end{itemize}

First, briefly explain your reasoning. Then assign a score from 0 to 10 according to the rubric below.

\textbf{Scoring Rubric:}
\begin{itemize}[leftmargin=0.3cm]
\item 0--1: Completely incorrect
\item 2--3: Wrong answer but loosely related
\item 4--6: Partial match or buried in irrelevant info
\item 7--8: Mostly correct with minor redundant details
\item 9--10: Fully correct and precise match
\end{itemize}

\textbf{Input Format:} \\
\textbf{Expected Answer:} \{reference answer(s)\} \\
\textbf{Actual Answer:} \{model-generated answer\}

\textbf{Output Format (JSON):} \\
\{ "reasoning": "...", "score": <0-10> \}
\end{promptbox}

\end{document}